\documentclass{article}
\usepackage{PRIMEarxiv}
\usepackage{microtype}
\usepackage{graphicx}
\usepackage{booktabs} % for professional tables
\usepackage{gensymb}
\usepackage{hyperref}

\usepackage{PRIMEarxiv}

\usepackage{amsmath}
\usepackage{amssymb}
\usepackage{mathtools}
\usepackage{amsthm}

\def\fighome{figures}

\usepackage{lineno,microtype}

\usepackage{xspace}
\usepackage{wrapfig}

\usepackage{graphicx,float,pgfplots,wrapfig,sidecap,lipsum}
\usepackage{tabularx}
\usepackage{booktabs}
\usepackage{paralist}
\usepackage[T1]{fontenc}
\usepackage{amsfonts}
\usepackage{amsthm}
\newtheorem{theorem}{Theorem}
\newtheorem*{theorem*}{Theorem}
\usepackage{amsmath}
\usepackage{amssymb}
\usepackage{mathrsfs}
\usepackage{xcolor}
\definecolor{lightblue}{RGB}{240,245,255}
\definecolor{darkblue}{RGB}{40,40,85}
\usepackage{listings}
\usepackage{enumerate}
\usepackage{enumitem}
\usepackage{mathtools} 

\usepackage{tablefootnote}
\usepackage{caption}
\usepackage{subcaption}
\usepackage[normalem]{ulem}
\usepackage{psfrag}
\usepackage{tikz}
\usetikzlibrary{fit}
\usetikzlibrary{calc,shapes}
\usetikzlibrary{decorations.pathmorphing} % noisy shapes
\usetikzlibrary{fit}					% fitting shapes to coordinates
\usetikzlibrary{backgrounds}	% drawing the background after the foreground

\usepackage{bm}
\usepackage{cleveref}

\usepackage{stmaryrd}
\usepackage[utf8]{inputenc}
\usepackage[english]{babel}

\usepackage{multirow}
\usepackage{tabularx}
\usepackage{makecell} %\thread

\usepackage{graphicx,float,pgfplots,wrapfig,sidecap,lipsum}
\pgfplotsset{compat=1.17}

\usepackage{tikz}
\usetikzlibrary{calc,trees,shadows,positioning,arrows,chains,shapes.geometric,decorations.pathreplacing,decorations.pathmorphing,shapes,
matrix,shapes.symbols,patterns,intersections,fit}
\newcommand{\tensor}[1]{\bm{\mathcal{#1}}}
\newcommand{\mytensor}[1]{\bm{\mathcal{#1}}}
\newcommand{\mymatrix}[1]{\bm{#1}}
\newcommand{\myvector}[1]{\bm{#1}}

\newcommand{\matrixSup}[2]{\bm{#1}^{(#2)}}

\newcommand{\tensorSub}[2]{\bm{\mathcal{#1}}_{#2}}
\newcommand{\tensorInd}[3]{\bm{\mathcal{#1}}^{(#2)}_{#3}}
\newcommand{\tensorSup}[2]{\bm{\mathcal{#1}}^{(#2)}}

\newcommand{\slice}{\bm{:}}

\ifdefined\R
\else
\newcommand\R{\mathbb{R}}
\fi

\def\rd{\mathrm{rd}}
\def\th{\mathrm{th}}

\newcommand{\einsum}{\textsf{einsum}\xspace}
\newcommand{\conveinsum}{\textsf{conv\_einsum}\xspace}

\newcommand{\convNd}{\textsf{convNd}\xspace}
\newcommand{\convOne}{\textsf{conv1d}\xspace}
\newcommand{\convTwo}{\textsf{conv2d}\xspace}

\newcommand{\pytorch}{\textsf{PyTorch}\xspace}
\newcommand{\numpy}{\textsf{NumPy}\xspace}
\newcommand{\tensorflow}{\textsf{TensorFlow}\xspace}

\newcommand{\opteinsum}{\textsf{opt-einsum}\xspace}
\newcommand{\netcon}{\textsf{netcon}\xspace}

\newcommand{\opentnn}{\textsf{OpenTNN}\xspace}

\title{\texttt{conv\_einsum}: A Framework for Representation and Fast Evaluation of Multilinear Operations in Convolutional Tensorial Neural Networks
}

\author{
  Tahseen Rabbani\thanks{Equal contribution.}\hspace{0.1em},\hspace{0.2em} Jiahao Su$^*$, Xiaoyu Liu, David Chan, Geoffrey Sangston, Furong Huang \\
  Department of Computer Science\\
  University of Maryland\\
  \texttt{\{trabbani,jiahaosu,xliu1231,dhchan,gsangsto,furongh\}@umd.edu} \\
}

\usepackage{tikz}
\usetikzlibrary{external}
\begin{document}
\maketitle
\begin{abstract}
Modern ConvNets continue to achieve state-of-the-art results over a vast array of vision and image classification tasks, but at the cost of increasing parameters. One strategy for compactifying a network without sacrificing much expressive power is to reshape it into a tensorial neural network (TNN), which is a higher-order tensorization of its layers, followed by a factorization, such as a CP-decomposition, which strips a weight down to its critical basis components. Passes through TNNs can be represented as sequences of multilinear operations (MLOs), where the evaluation path can greatly affect the number of floating point operations (FLOPs) incurred. While functions such as the popular \einsum can evaluate simple MLOs such as contractions, existing implementations cannot process multi-way convolutions, resulting in scant assessments of how optimal evaluation paths through tensorized convolutional layers can improve training speed. In this paper, we develop a unifying framework for representing tensorial convolution layers as \einsum-like strings and a meta-algorithm \conveinsum which is able to evaluate these strings in a FLOPs-minimizing manner. Comprehensive experiments, using our open-source implementation, over a wide range of models, tensor decompositions, and diverse tasks, demonstrate that \conveinsum significantly increases both computational and memory-efficiency of convolutional TNNs.
\end{abstract}

\section{Introduction}
\label{sec:introduction}

Modern neural networks are both expressive and accurate over a wide variety of learning and classification problems, but at the cost of increased width and depth. State-of-the-art convolutional models, for example, can contain up to several billion parameters \cite{khan2020survey,szegedy2015going,brown2020language}. The size and training costs of such networks are at odds with a rapidly emerging industrial and academic interest in performing learning tasks on low-fidelity hosts such as IoT and mobile devices \cite{mcmahan2017communication,li2018learning,kairouz2019advances}. An increasingly popular approach for generating compact yet expressive models is to use Tensorial Neural Networks (TNNs) \cite{lebedev2015speeding,kossaifi2017tensor,kossaifi2020tensor,su2018tensorial}, which factorize each layer’s weight tensor into several smaller factors. As a result, each TNN layer is a multilinear operation of its input and weight factors.

One method by which a TNN can arise is reshaping a network weight into a higher order tensor. These reshaped weights are then decomposed into factorized forms, such as canonical polyadic (CP), tensor train (TT), and Tucker (TK) decompositions. Factorization of a reshaped tensor is an efficient way of representing underlying properties such as periodicity and modularity invariances/similarities, which often exist in neural network models \cite{lebedev2015speeding,garipov2016ultimate,su2018tensorial,wang2018wide}, and preserved in the factorization. In addition to their lighter weight, TNNs can preserve the same predictive power level over several popular backbone networks and tasks. 

However, in contrast to the availability of high performance libraries for convolutional neural networks (CNNs), e.g., NVIDIA’s cuDNN for GPUs or Intel’s MKL for CPUs), solutions for efficient evaluation of multilinear operations (MLOs) through their tensor-decomposed counterparts are relatively unavailable. Emphasis in existing literature and implementations has been placed on speed-up via compression, i.e., reducing the number of MLOs required by tensorial convolutional kernels by reducing the number of their parameters via rank-reduction. Few works have focused on the reduction of convolutional MLOs when the size of the layer is fixed. Since the convolution operator is fundamental in many cutting-edge architectures and highly optimized by parallel computing libraries such as CUDA, support for this multilinear operation in TNN training, would benefit the tensor network community.

%and fine-tuned decomposition for shorter, stabler training

%This is especially useful for low-rank, higher-order reshaped weights since we can capture abundant structural information without much representation redundancy. \Cref{fig:periodicity} depicts this scenario using the example of a vector displaying periodicity.

% Figure for reshaping
%\input{\texhome/fig_reshaping}
%Since the convolution operator is fundamental in many cutting-edge architectures and highly optimized by parallel computing libraries such as CUDA, support for this multilinear operation in TNN training, would benefit the tensor network community.

In this work: \textbf{(1)} We introduce a framework for representation and optimal evaluation  order of multilinear floating point operations (FLOPs) through convolutional tensor layers. %\textcolor{blue}{Given a backbone \textcolor{blue}{tensorial neural network}, \autotnn can compress the tensorial weights, and users may make use of \conveinsum to describe the tensor operations as \einsum strings (including convolutions) to conduct end-to-end training on the resulting TNN.}
\textbf{(2)} Using our framework, one can represent a tensorial forward or backward computation as a \textit{generalized} \einsum graph/sequence and submit it to our novel meta-algorithm \conveinsum for FLOPs-minimized evaluation. Here, ``generalized'' means including convolutions, an operation not supported by any existing \einsum implementation. To determine an optimal evaluation path, we develop an extension of the \netcon algorithm \cite{pfeifer2014faster} to support the expense of convolutions in its underlying cost model. \textbf{(3)} We present theory and a comprehensive set of experiments which demonstrate significant improvements in computational and memory efficiency by using \conveinsum to train TNNs. \textbf{(4)} Our code is available as an open-source library for training tensorized ResNets under a variety of compression rates and factorizations.

\section{Tensor Operations and \einsum}
\label{sec:preliminary}

In this section, we outline multi-linear operations (MLOs) common to TNNs. In brief, MLOs are nested summations over one or more indices/modes of a tensor and described in full generality in Appendix \ref{app-sub:multiops}. Many of these operations are systematically expressible and computable via the popular notational framework and function \einsum, first introduced by the Python library \numpy~\cite{harris2020array}.
Therefore, we will first review the most important tensor network operations and their corresponding \einsum representations. We then formally introduce TNNs.

\textbf{Notations.}
We use lower case letters (e.g., $\myvector{v}$) to denote vectors, 
upper case letters (e.g., $\mymatrix{M}$) denote matrices, 
and curly letters (e.g., $\mytensor{T}$) denote tensors.
For a tensor $\mytensor{T} \in \R^{I_1 \times I_2 \times \cdots \times I_N}$, we refer to a specific entry by the subscript notation $T_{i_1 i_2 \dots i_N}$, where $1 \leq i_n \leq I_n$ for $1 \leq n \leq N$.
We refer to $N$ as the order, a single index $n$ as a mode, and the magnitude of a mode $I_n$ as dimension size.
For example, a tensor $\mytensor{T} \in \R^{3 \times 4 \times 5}$ is a $3^\rd$ order tensor that has dimension size $4$ at its second mode.

\subsection{Representations of Multilinear Operations}
\label{sub:multi-ops}

The \einsum function allows definitions of multilinear operations via string inputs. In this subsection, we highlight an example of a multilinear operation involving three primitive operations ({\em contraction}, {\em batch product}, {\em outer product}) in the language of \einsum.
% simultaneous operations between two tensors
Consider two $3^\rd$-order tensors $\tensorSup{T}{1} \in \R^{B \times C \times I}$, $\tensorSup{T}{2} \in \R^{A \times R \times T}$, and a multilinear operation between them: $\tensorSub{T}{b,i,j} = \sum_{c = 1}^{C}
\tensorInd{T}{1}{b,c,i} \tensorInd{T}{2}{b,c,j}$.

We can denote the operation above in \einsum as:
\begin{lstlisting}
T = einsum("bci,bcj->bij", T1, T2)
\end{lstlisting}
where the string in the quotation mark precisely specifies the operation, which is known as an \einsum string. In this string, the letter \textsf{"c"} indicates {\em contraction} since it appears in both inputs but not the output; the letter \textsf{"b"} symbolizes {\em batch product} since it appears in both inputs and the output; lastly, the letters \textsf{"i"} and \textsf{"j"} represent {\em outer product} as they each appears in one of the two inputs and they both appear in the output. %Note that any letter in an \einsum string falls into one of these three classes (except for the trivial case that a letter appearing in only one input or output). 
Detailed description of \einsum and these operations is provided in Appendix \ref{app-sub:multiops}.

\subsection{From \einsum to \conveinsum}
\label{sub:einsum}
While many popular libraries (e.g., \numpy, \tensorflow, \pytorch) implement \einsum, none directly support convolutions, despite the ubiquity of convolutions in modern neural networks.
We will generalize \einsum to a meta-function \conveinsum, 
which can process convolutions.

\textit{Tensor convolution} generalizes the convolution on vectors to higher-order tensors thereby extending the familiar linear convolution to a multi-linear operation.
Given two tensors $\tensorSup{T}{1} \in \R^{X \times B \times C}$ and $ \tensorSup{T}{2} \in \R^{L \times D \times E}$, we can define a convolution between the modes with dimensions $X$ and $L$. The operation returns a $5^\th$ order tensor $\mytensor{T} \in \R^{X^\prime \times B \times C \times D \times E}$, with its entries calculated as:
\begingroup
\abovedisplayskip=5pt
\belowdisplayskip=5pt
\begin{equation}
\label{eq:convolution}
\tensorSub{T}{\slice,b,c,d,e} = 
\tensorInd{T}{1}{\slice,b,c} \ast
\tensorInd{T}{2}{\slice,d,e},
\end{equation}
\endgroup
where $\ast$ denotes a convolution between two vectors. Note that the dimension sizes $X$ and $L$ can be different, and the output dimension size $X^\prime$ depends on the convolution type (e.g., a standard convolution yields $X^\prime = X + L - 1$).
We write \Cref{eq:convolution} in our proposed \conveinsum as
\begin{lstlisting}
T = conv_einsum("xbc,ade->xbcde|x", T1, T2)
\end{lstlisting}

In this scheme, the same letter "x" is used for different modes, even if their dimension sizes may differ. Furthermore, the placement "x" to the right of the  pipe-delimiter "$|$" indicates that \conveinsum performs convolution on the corresponding modes. 
We can use \conveinsum to represent multilinear operations on more than two inputs. Consider three tensors $\mytensor{X} \in \R^{B \times F \times S \times H \times W}$, $\tensorSup{K}{1} \in \R^{F \times G \times K \times K}$, $\tensorSup{K}{2} \in \R^{S \times T \times K \times K}$, and an operation
\begingroup
\abovedisplayskip=5pt
\belowdisplayskip=5pt
\begin{equation}
\tensorSub{Y}{b,g,t,\slice,\slice} = \sum_{f = 1}^{F} \sum_{s = 1}^{S} 
\tensorSub{X}{b,f,s,\slice,\slice} \ast \tensorInd{K}{1}{f,g,\slice,\slice} \ast \tensorInd{K}{2}{s,t,\slice,\slice} 
\end{equation}
which leads to an output tensor $\mytensor{Y} 
\in \R^{B \times G \times T \times H^\prime \times W^\prime}$. This is known as {\em interleaved group convolution}~\cite{zhang2017interleaved}. In \conveinsum, it writes as
\begin{lstlisting}
T = conv_einsum("bfshw,fghw,sthw->bgthw|hw",X, K1, K2)
\end{lstlisting}

\subsection{Compact Neural Networks via \conveinsum}
\label{sub:TNN}
 
In this section, we formulate various layers of compact neural networks in terms of \conveinsum. Expressions for many other layer types can be found in the Appendix.

\textbf{Standard convolution layer.} 
% We first review two of the most popular layer types, namely {\em fully-connected layers} and {\em 2D convolution layers}.
% % fully-connected layer
% {\bf (1)} A fully-connected layer is parameterized by a matrix $\mymatrix{W} \in \R^{T \times S}$, which maps a vector $\myvector{x} \in \R^{S}$ to an output vector $\myvector{y} \in \R^{T}$:
% \begin{lstlisting}
% Y = conv_einsum("bs,ts->bt", X, W)
% \end{lstlisting}
% \vspace{-1em}
% Since a neural network typically computes its inputs in mini-batches, the \einsum string contains an additional letter \textsf{"b"} to index examples in a mini-batch.
% 2D-convolutional layer
We review a popular layer type in neural networks --- the standard 2D-convolutional layer.
Such a layer is parameterized by a $4^\th$ order tensor $\mytensor{W} \in \R^{T \times S \times H \times W}$, which maps a $3^\rd$ order tensor $\mytensor{X} \in \R^{S \times H^\prime \times W^\prime}$ to a $3^\rd$ order tensor $\mytensor{Y} \in \R^{T \times H^\prime \times W^\prime}$:
\begin{lstlisting}
Y = conv_einsum("bshw,tshw->bthw|hw", X, W)
\end{lstlisting}
Note that a neural network typically computes its inputs in mini-batches, so the \conveinsum string contains an additional letter \texttt{"b"} to index examples in a mini-batch. Since convolutional layers include fully-connected layers as a special case when $H = W = 1$, we focus on designs of convolutional layers for the remainder of this subsection.

% figure for a convolution layer
% \input{\texhome/figure_conv_layer}

\textbf{Tensorial Neural Networks.}
Convolutional layers motivate the importance and usage of TNNs since their structure has been shown to benefit from reshaping and tensorial decomposition. Numerous works propose to design {\em tensorial layers} where the (reshaped) convolution kernel $\mytensor{W}$ is factorized using tensor decompositions~\cite{lebedev2015speeding,kim2015compression,garipov2016ultimate,wang2017tensor,su2018tensorial}. Our proposed \conveinsum can handle these types of designs. 
Here, we present two representatives of efficient tensorial convolutional layer designs based on the CP decomposition~\cite{kolda2009tensor}.

% CP-convolutional layer
{\bf (1)} In a {\em CP convolutional layer}~\cite{lebedev2015speeding}, the kernel $\mytensor{W}$ is factorized into $4$ factors $\matrixSup{W}{1} \in \R^{R \times T}$, $\matrixSup{W}{2} \in \R^{R \times S}$, $\matrixSup{W}{3} \in \R^{R \times H}$, $\matrixSup{W}{4} \in \R^{R \times W}$ such that
\begin{lstlisting}
W = conv_einsum("rt,rs,rh,rw->tshw", W1, W2, W3, W4)
\end{lstlisting}

Plugging this decomposition into the 2D-convolutional layer, we obtain the following \conveinsum output string:
\begin{lstlisting}
Y = conv_einsum("bshw,rt,rs,rh,rw->bthw|hw", X, W1, W2, W3, W4)
\end{lstlisting}

% reshaped CP convolutional layer
{\bf (2)} For a {\em reshaped CP convolutional layer}~\cite{su2018tensorial}, the convolution kernel $\mytensor{W} \in \R^{T \times S \times H \times W}$ is first reshaped into a higher order tensor $\mytensor{\overline{W}} \in \R^{T_1 \cdots \times T_M \times S_1 \cdots \times S_M \times H \times W}$ such that $T = \prod_{m = 1}^{M} T_m$, $S = \prod_{m = 1}^{M} S_m$, and then factorized into $(m + 1)$ tensors $\matrixSup{W}{m} \in \R^{R \times T_m \times S_m}$ with $\tensorSup{W}{0} \in \R^{R \times H \times W}$. For example, when $M = 3$:
\begin{lstlisting}
W = conv_einsum("r(t1)(s1),r(t2)(s2),r(t3)(s3),rhw"->(t3)(t2)(t1)(s3)(s2)(s1)", W1, W2, W3, W0)
\end{lstlisting}
We can write the layer's \conveinsum string as:
\begin{lstlisting}
Y = conv_einsum("b(s1)(s2)(s3)hw,r(t1)(s1),r(t2)(s2),r(t3)(s3),rhw->n(t1)(t2)(t3)hw",X, W1, W2, W3, W0)
\end{lstlisting}

For both layers, $R$ is the {\em rank} of the CP decomposition, which controls the number of parameters (i.e., compression rate) of the layer.

A number of other works present alternative designs for efficient convolutional layers such as the \emph{interleaved group convolution}~\cite{zhang2017interleaved} and \emph{separable depth-wise convolution}~\cite{chollet2017xception}. As discussed in Appendix \ref{app:efficientNN}, our proposed \conveinsum also covers these specialized designs. We also refer the reader to \cite{su2018tensorial,hayashi2019exploring} for more examples.

\section{Algorithms}
\label{sec:algorithms}

In the previous section, we presented how \conveinsum represents general multi-linear operations in (compact) neural networks.
In this section, we develop a suite of algorithms to implement the \conveinsum function efficiently. We organize this section as follows: 
\textbf{(1)} First, we develop an algorithm to reduce a \conveinsum function with two inputs to a collection  of atomic \pytorch operations, which allows us to reuse GPU-optimized functions in \pytorch to complete the computation. 
\textbf{(2)} Secondly, we derive an optimal sequencer which automatically decomposes a \conveinsum function with an arbitrary number of inputs into a FLOPs-minimal sequence of 2-input \conveinsum functions;
\textbf{(3)} Lastly, we describe how gradient checkpointing \cite{chen2016training} can be used to reduce the memory overhead in the backpropagation phase.

\subsection{Atomic Operations}
\label{sub:algorithms-atomic}

We can represent a \conveinsum function with two inputs via GPU-optimized \pytorch functions, specifically \einsum and \convNd (e.g., \convOne, \convTwo).
%If there is no letter for convolution, we can simply use \einsum function.
In particular, we will show that any \conveinsum function with convolution can be realized via \convNd. To understand why such a transformation is possible, we first analyze the \conveinsum string for the \convOne function:
\begin{lstlisting}
Y = conv_einsum("bsh,tsh->bth|h", X, W)
\end{lstlisting}

where \textsf{"t"} stands for the output channel, \textsf{"s"} the input channel, \textsf{"h"} the length of features/filters, and \textsf{"b"} the batch size. Now, we can categorize these letters in terms of primitive operations.
\textbf{(1)} The letter \textsf{"h"} is a convolution, 
appearing in both inputs and the output;
\textbf{(2)} The letter \textsf{"s"} is a contraction, 
appearing in both inputs but not the output;
\textbf{(3a)} The letter \textsf{"t"} is an outer product,
appearing in the first input and the output;
\textbf{(3b)} The letter \textsf{"b"} is another outer product,
appearing in the second input and the output.

The \convOne function covers almost all mixtures of compound multi-linear operations in which each operation type appears at most once, which we refer to as an \textit{atomic operation}, except two cases:
\textbf{(4)} A batch product that appears in both inputs and the output; and
\textbf{(5)} A self-contraction that appears in only one input.
Fortunately, these two edge cases can be easily addressed.
For \textbf{(4)}, the function \convOne supports a group-convolution option, which effectively extends to:

\begin{lstlisting}
Y = conv_einsum("gtsh,bgsh->bgth|h", X, W)
\end{lstlisting}

where \textsf{"g"} stands for the filter group. In terms of tensor operations, it is a batch product, which appears in both inputs and the output. For \textbf{(5)}, such a letter can be eliminated by summing over the corresponding index in pre-processing.

In summary, \textbf{(1)-(5)} cover all possible types of tensor operations; the function \convOne covers \textbf{(1)-(4)} and \textbf{(5)} can be addressed via a pre-processing. Paired with index-swapping operations, \convOne can realize any \conveinsum string where each operation type appears only once.

% Discussion of multiple letters
\textbf{Multiple letters with the same operation type.} 
Now we will address the scenario where there are multiple letters with the same operation type. For example, if there are two different letters in a \conveinsum string designated for convolution, we can use \convTwo instead of \convOne. Notice that \convTwo realizes a \conveinsum such as:
\begin{lstlisting}
Y = conv_einsum("gtshw,bgshw->bgthw|h,w", X, W)
\end{lstlisting}
where \textsf{"g"}  stands for the filter group and \textsf{"h"}/\textsf{"w"}  represent height/width of the filters/features respectively. In principle, we can use a \convNd function to compute a \conveinsum function with $N$ letters for convolution (though \convNd for $N \geq 4$ requires custom implementation).

For non-convolution letters, all letters with the same type can be merged into one letter in the preprocessing (i.e., the corresponding modes are reshaped into one compound mode), and the letter is converted back to multiple letters in the post-processing step (i.e., the compound mode is reshaped into its corresponding modes).

% figure for optimal sequencer
\begin{figure}[!htbp]
\begin{minipage}[!htbp]{\linewidth}
\begin{centering}
\begin{lstlisting}
A=np.random.rand(4,7,9)
B=np.random.rand(10,5)
C=np.random.rand(5,4,2)
D=np.random.rand(6,8,9,2)
path_info = conv_einsum.contract_path("ijk,jl,lmq, njpq->ijknp|j", A, B, C, D)
print(path_info[1])
\end{lstlisting}
\subcaption{\textbf{Tensor sequence generation.} We analyze a sequence over a collection of tensors $\tensor{A}, \tensor{B}, \tensor{C}$, $\tensor{D}$, involving contraction, convolution, and batch product (Python). We store the optimal sequence in a string array \texttt{path\_info}.}
\label{fig:cvoutput-a}
\resizebox{\textwidth}{!}{
    \includegraphics{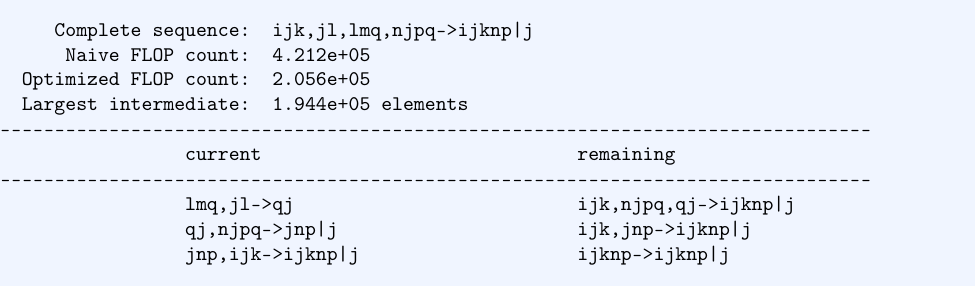}
    }
\subcaption{\textbf{An optimal sequence of paths.} Visualization (via \opteinsum)  of the optimal sequence of paths for the \conveinsum string submitted in \Cref{fig:cvoutput-a}, with our modified support for convolution symbols. We are also presented with information about the naive left-to-right cost vs the cost of the suggested path.}
\label{fig:cvoutput-b}
\end{centering}
\end{minipage}
\caption{\textbf{\conveinsum sample code.} The figure depicts the generation and analysis of a set of tensors coalesced into one tensor sequence (Python).}
\label{fig:cvoutput}
\end{figure}

\begin{figure*}[!ht]
\begin{centering}
 \includegraphics[width=0.9\textwidth]{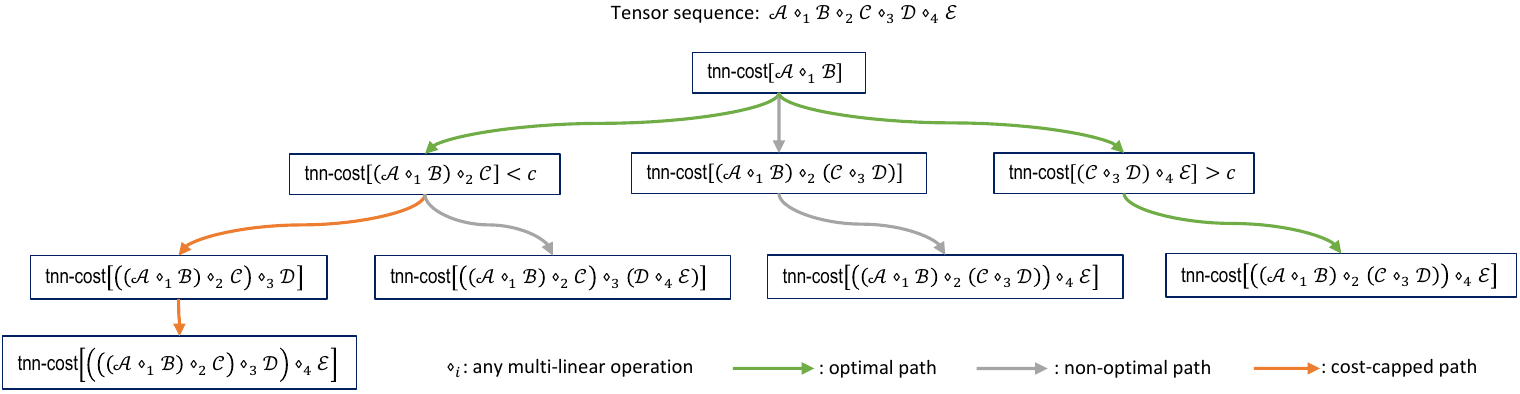}
 \vspace{-0.5em}
 \caption{\textbf{Optimal sequencer example}. \conveinsum deploys the optimal sequencer to analyze the path tree of an abstract tensor sequence $\tensor{A} \circ_1 \tensor{B} \circ_2 \tensor{C} \circ_3 \tensor{D} \circ_4 \tensor{E}$, where $\circ_i$ for $1\leq i \leq 4$ is any collection of multi-linear operations, including convolutions, batch products, contractions, and outer products. %The optimal sequencer, which is a tree traversal strategy, fuses \netcon and our \textsf{tnn-cost} function. 
 The green path indicates the optimal path and the orange path indicates a path which satisfies a user-specified cost cap $c$ at each node.
 }
 \label{fig:opt-seq}
\end{centering}
\end{figure*}

\subsection{Optimal Sequencer}
\label{sub:algorithms-sequencer}

The \conveinsum strategy for optimal sequence discovery will work with an extensive variety of decompositions and with any existing open-source einsum implementation such as \texttt{numpy.einsum} (\texttt{Numpy}), \texttt{tf.einsum} (TensorFlow), or \opteinsum \cite{daniel2018opt}, which sits on top of \texttt{NumPy}. These functions/libraries can natively handle determining the FLOPs-optimal evaluation order of tensor networks involving (non-convolutional) operations via the \netcon algorithm \cite{pfeifer2014faster}. We add support for convolutions inside the \netcon algorithm; we refer to the extended algorithm as the \textit{optimal sequencer} within our framework. We forked and modified the \opteinsum library to handle convolutions as a proof of concept. We present a high-level overview of the optimal sequencer which can be used for a \texttt{numpy.einsum} or \texttt{tf.einsum} implementation.

\textbf{Better than left-to-right evaluation.} \netcon was designed to handle general contraction sequences in tensor networks. For example, for $\tensor{A} \in \R^{I \times J \times K}, \tensor{B} \in \R^{J \times L}, \tensor{C} \in \R^{L \times M}$, one might be interested in the optimal order of evaluation of the tensor 
$\tensor{T} = \sum_{l = 1}^{L} \sum_{j = 1}^{J}
\tensorSub{A}{\slice, j, \slice} \otimes \tensorSub{B}{j,l} \otimes \tensorSub{C}{l, \slice}$.
%\begingroup
%\abovedisplayskip=5pt
%\belowdisplayskip=5pt
%\begin{equation}
%\label{eq:net-consequence}
%\tensor{T} = \sum_{l = 1}^{L} \sum_{j = 1}^{J}
%\tensorSub{A}{\slice, j, \slice} \otimes \tensorSub{B}{j,l} \otimes \tensorSub{C}%{l, \slice}.
%\end{equation}
%\endgroup
Let us momentarily suppress the index and summation notation of this equation, i.e., let $(AB) \triangleq \sum_{j=0}^{J-1}\mathcal{A}_{:,j,:}\mathcal{B}_{j,l}$, $(BC)\triangleq \sum_{l=0}^{L-1}\mathcal{B}_{j,l}\mathcal{C}_{l,:}$. The possible paths we may take to arrive at $\tensor{T}$ include $(AB)\rightarrow (AB)C$ (the so-called \textbf{left-to-right} path), or $(BC)\rightarrow A(BC)$. Each of these paths has a predictable \textit{contraction cost}, or number of multiplications/additions (FLOPs), dependent on the dimensions of the tensor modes involved in each intermediate product. \netcon is designed to efficiently traverse such cost-path trees and determine the FLOPs-optimal path in a fast manner.

In particular, $\netcon$ is capable of handling all types of MLO sequences we have described except for those containing convolutions. For example, consider the tensor $\tensor{T}_{p,:,q,r,t}=\sum_{n=1}^N \tensor{B}_{n,p}\bigl(\tensor{A}_{n,:,r} \ast \tensor{C}_{:,q}\bigr)\tensor{D}_{r,t}$ which contains a convolution. 

Our optimal sequencer extends the \netcon paradigm to handle convolutions by replacing the contraction cost function with a more general \texttt{tnn-cost} function, which adds the cost (FLOPs-wise) of the convolutions (if present) within an intermediate product at each node. Further details of the logic is deferred to Appendix \ref{app:algorithms}. \Cref{fig:opt-seq} depicts the optimal sequencer analyzing an abstract tensor sequence. Since \einsum-like evaluation of tensorial convolutional layers was previously not possible, training speed up via optimal path evaluation of MLOs containing convolutions was difficult to explore. We now present a pair of theorems which assert the existence of cheaper-than-naive evaluation path for an RCP or RTK convolutional layer. 
\begin{theorem}[CP reduction]
\label{flop-reduce}
Let $\tensor{X}\in\mathbb{R}^{B \times S\times H' \times W'}$ be the input to a reshaped CP (RCP) convolutional kernel $\mytensor{\overline{W}} \in \R^{T_1 \cdots \times T_M \times S_1 \cdots \times S_M \times H \times W}$ such that $T = \prod_{m = 1}^{M} T_m$, $S = \prod_{m = 1}^{M} S_m$ are factored into $(m + 1)$ rank-$R$ CP factor tensors $\matrixSup{W}{m} \in \R^{R \times T_m \times S_m}$ with $\tensorSup{W}{0} \in \R^{R \times H \times W}$. Assume $H'\gg H$ and $W'\gg W$ are large; in particular $SH'W'>aHW$ and $BH'W'>aS$ for some constant $a\geq 1$. Furthermore, let $R\geq S$. Then the forward pass through the RCP kernel (in the syntax of \conveinsum), 
\begin{lstlisting}[mathescape=true]
Y=conv_einsum("b(s1)(s2)(s3)$\cdots$(sM)hw,r(t1)(s1),r(t2)(s2),r(t3)(s3),$\dots$, r(tM)(sM),rhw->r(t1)(t2)(t3)$\cdots$(tM)hw|hw",X, W1, W2, W3, $\dots$, WM, W0)
\end{lstlisting}
has a pairwise evaluation path which costs less FLOPs than the naive left-to-right evaluation,
 \begin{lstlisting}[mathescape=true]
Y=conv_einsum("br(t1)(t2)(t3)$\cdots$(tM)hw, rhw -> br(t1)(t2)(t3)$\cdots$(tM)hw| hw", YM, W0)
 \end{lstlisting}
 where
 \begin{lstlisting}[mathescape=true]
 Ym=einsum("r(s1)(s2)$\cdots$(s(m-1))(t1)(t2)$\cdots$(t(m-1)),r(sm)(tm)->r(s1)(s2)$\cdots$(sm)(t1)(t2)$\cdots$ (tm), Y(m-1), Wm) 
 \end{lstlisting}
 for $1\leq m \leq M$, noting that for tensor object \texttt{YM}, its mode symbols \texttt{h}, \texttt{w} correspond to dimensions $H',W'$, and \texttt{b} corresponds to an arbitrary batch size.
\end{theorem}

\begin{theorem}[Tucker reduction]
\label{flop-reduce-tucker}
Let $\tensor{X}\in\mathbb{R}^{B \times S\times H' \times W'}$ be the input to a reshaped Tucker (RTK) convolutional kernel $\mytensor{\overline{W}} \in \R^{T_1 \cdots \times T_M \times S_1 \cdots \times S_M \times H \times W}$ such that $T = \prod_{m = 1}^{M} T_m$, $S = \prod_{m = 1}^{M} S_m$ are factored into $(m + 1)$ rank-$R_m$ CP factor tensors $\matrixSup{W}{m} \in \R^{R_m \times T_m \times S_m}$ with $\tensorSup{W}{0} \in \R^{R \times H \times W}$ and $\tensorSup{C} \in \R^{R_0 \times R_1 \times \cdots \times R_M}$. Assume $H'\gg H$ and $W'\gg W$ are large; in particular $SH'W'>aHW$ and $BH'W'>aS$ for some constant $a\geq 1$. Furthermore, let $\prod_{i=1}^M R_m\geq S$. Then the forward pass through the RCT kernel (in the syntax of \conveinsum), 
\begin{lstlisting}[mathescape=true]
Y=conv_einsum("b(s1)(s2)(s3)$\cdots$(sM)hw,(r1)(t1)(s1),(r2)(t2)(s2),(r3)(t3)(s3),$\dots$, (rM)(tM)(sM),(r0)hw, (r0)(r1)(r2)$\cdots$(rM)->b(t1)(t2)(t3)$\cdots$(tM)hw|hw",X, W1, W2, W3, $\dots$, WM, W0, C)
\end{lstlisting}
has a pairwise evaluation path which costs less FLOPs than the naive left-to-right evaluation,
 \begin{lstlisting}[mathescape=true]
Y=conv_einsum("br(t1)(t2)(t3)$\cdots$(tM)hw, rhw -> br(t1)(t2)(t3)$\cdots$(tM)hw| hw", YM, W0)
 \end{lstlisting}
 where
 \begin{lstlisting}[mathescape=true]
 Ym=einsum("(r(m-1))(s1)(s2)$\cdots$(s(m-1))(t1)(t2)$\cdots$(t(m-1)),(rm)(sm)(tm)->r(s1)(s2)$\cdots$(sm)(t1)(t2)$\cdots$ (tm), Y(m-1), Wm) 
 \end{lstlisting}
 for $1\leq m \leq M$, noting that for tensor object \texttt{YM}, its mode symbols \texttt{h}, \texttt{w} correspond to dimensions $H',W'$, and \texttt{b} corresponds to an arbitrary batch size.
\end{theorem}
\textit{Proof sketches.} The full proof is deferred to the Appendix. It uses an example path which sequentially reconstructs $\mytensor{\overline{W}}$ avoids the addition of a large $\mathcal{O}(H'W')$ cost to all intermediates, which occurs in the CP and Tucker naive evaluations.
%\begin{equation}
%\mathcal{O}\bigl(H'W'\sum S_i\sum T_i +HWTRH'W'\bigr)  
%\end{equation}

\subsection{Checkpointing}
\label{sub:algorithms-checkpointing}

Since we evaluate a tensor network in a pairwise manner, computing a tensor network with $N$ inputs leads to $(N - 1)$ intermediate results. If we use an autograd-like function, we will need to save these intermediates in memory, causing high memory overhead.
To avoid storing intermediate products, one can use gradient checkpointing~\cite{chen2016training}
which recomputes the gradient during the backward pass rather than saving all intermediate results in memory. Normally, in the forward pass, the model caches all values of the activation neurons and reuses them in the backwards pass calculation, which gradient checkpointing avoids.

\section{Related works}
\label{sec:related}
\textbf{Tensor networks.}
Tensor networks are widely used in quantum physics~\cite{orus2014practical}, 
numerical analysis~\cite{grasedyck2013literature}, and machine learning~\cite{cichocki2016tensor,cichocki2017tensor}. 
In neural networks, \cite{cohen2016convolutional} and \cite{khrulkov2018expressive} 
use tensor networks to prove the expressive power of convolutional and recurrent neural networks. Recently, \cite{hayashi2019exploring} combine tensor networks with genetic algorithms to search for efficient layer designs.
Traditional tensor networks do not support {\em convolutions} in their operations, which is essential in convolutional neural networks. However, in recent years, so-called quantum convolutional neural networks (QCNNs) used for tasks such as quantum phase classification have been shown to share the same circuit structure as multiscale entanglement renormalization ansatzes (MERAs), which are classical tensor networks used to simulate many-body systems. Interestingly enough, \cite{pfeifer2014faster} used \netcon to optimize multilinear operations through MERAs, but did not consider convolutions. 

\textbf{Low-rank factorization.} 
Various types of low-rank factorization have been proposed to reduce the number of parameters in linear layers.
% stage 1: flatten into matrix + (dictionary learning / singular value decomposition)
Pioneering works proposed to flatten/unfold the parameters in convolutional layers into matrices (known as {\em matricization}), followed by (sparse) dictionary learning or matrix decomposition~\cite{jaderberg2014speeding,denton2014exploiting,zhang2015efficient}.
% (2) Direct decomposition
Subsequently, \cite{lebedev2015speeding} and \cite{kim2015compression} showed that it is possible
to compress the parameters directly by standard tensor decompositions
(in particular, CP or Tucker~\cite{kolda2009tensor}). These decompositions, particularly of convolutional kernels, can occasionally result in degenerate components with high intensity/Frobenius norm, so error-corrective mechanisms have since been introduced \cite{phan2020stable}.
% (3) Tensorization
Further groundbreaking work ~\cite{novikov2015tensorizing,garipov2016ultimate} demonstrated that the low-order weights can be efficiently compressed by the tensor-train decomposition~\cite{oseledets2011tensor} by first reshaping the parameters into higher-order tensors.
% (4) Extension to recurrent neural network
This paradigm was later extended in two directions: (1) the tensor-train decomposition is used to compress LSTM/GRU layers in recurrent neural networks (RNNs)~\cite{yang2017tensor} and higher-order RNNs ~\cite{yu2017long,su2020convolutional}; and (2) other decompositions are explored for better compression, 
such as the tensor-ring decomposition~\cite{zhao2016tensor} and block-term decomposition~\cite{ye2020block}.

%\textbf{Other compression methods.} 
%TNNs belongs to the large family of low-rank approximation methods, complementary to other compression techniques such as quantization and pruning. Many papers have justified that combining these different lines of the works may render more competitive model compression. For instance, \cite{lee2021qttnet} verify that TNNs with quantization achieve SOTA results on 3D tasks, and \cite{goyal2019compression} demonstrate that TNNs with pruning obtain SOTA image classification. This work, however, is chiefly concerned with the automated development and benchmarking of TNNs without any other compression methods.

%but for future work, we will investigate incorporating quantization, pruning, and knowledge distillation techniques directly into \autotnn.}

\textbf{Existing Libraries/Algorithms.}
Various libraries support tensor operations. \textbf{(1) Pytorch}~\cite{paszke2019pytorch} supports specialized tensor operations commonly used in neural networks, including various convolutional layers. However, its \texttt{torch.einsum} cannot evaluate convolutional MLOs. \textbf{(2) TensorLy}~\cite{kossaifi2019tensorly} supports tensorization of common layer types across various platforms, including \pytorch and \tensorflow. \textbf{(3) NumPy}~\cite{harris2020array} is a general computation library that has an optimal sequencer in its \einsum function, but it does not support convolutions.
 \textbf{(4) OptEinsum}~\cite{daniel2018opt} interfaces with most other \einsum implementations and can further accelerate their computation via specialized BLAS routines. Due to its native implementation of \netcon and path visualization capabilities, we develop \conveinsum inside this library. 
\textbf{(5) Einops}~\cite{einops} extends the \einsum functionality to GPUs.
\textbf{(6) Gnetcon}~\cite{reustle2020fast} attempts to extend \einsum to convolutions but does not support multi-way convolution. \textbf{(7) TedNet} \cite{pan2022tednet} is a toolkit for computing low-rank tensor decompositions of common layer types.  \textbf{(8) Tensor Comprehensions (TC)} \cite{vasilache2018tensor} provides a language for expression for MLOs, including 2D convolutions, but it cannot easily process higher order convolutions. 

\section{Experiments}
\label{sec:experiments}

In this section, we demonstrate that \conveinsum greatly improves computational and memory efficiency of convolutional TNN training over a variety of tasks.

\textbf{Tasks.} 
%We test \conveinsum on a range of tasks under different network compression rates:
% (1) video classification (VC)
\textbf{(1)} A classic two-stream convolutional neural network \cite{simonyan2014two} is used for a video classification task, trained on the UCF-101 data set \cite{Soomro2012UCF101AD}. ResNet-101 \cite{he2016deep} was chosen as the ConvNet for both the spatial and temporal streams, pre-trained on ImageNet~\cite{deng2009imagenet}. The two-stream network is adapted from \cite{huang2019twostream}.
% (2) automatic speech recognition (ASR)
\textbf{(2)} An Automatic Speech Recognition task using the Conformer architecture \cite{gulati2020conformer}, which incorporates convolution modules between the attention modules and the feed-forward modules of a Transformer model~\cite{vaswani2017attention}. The model is trained from scratch with random normal initialization on the LibriSpeech dataset~\cite{panayotov2015librispeech}.
% (3) image classification (IC)
\textbf{(3)} Image classification tasks with model trained from scratch on CIFAR10~\cite{krizhevsky2009learning} and ImageNet datasets using ResNet-34.

Weight decay is set to $5\times 10^{-4}$, momentum is set to $0.9$, and learning rate is initialized as $0.05$ with decay rate of $0.5$ every 30 epochs. An NVIDIA GeForce RTX 2080Ti is used for all experiments. All training is implemented in \pytorch. For all RCP experiments, $M=3$, and only the input and output channel modes are reshaped to a third order tensor. 

\textbf{TensorFlow vs PyTorch.} Although all our training is done in PyTorch, we can reduce higher-order convolutions to atomic conv1d and conv2d operations in both PyTorch \textit{and} TensorFlow per Section \ref{tab:primitive-operations}. Computation of MLOs in both PyTorch and TensorFlow ultimately reduces to the same CUDA backend, so we expect performance comparisons to be similar. Furthermore, the number of FLOPs required to compute a fixed layer type is independent of the backend library and purely a function of the tensor dimensions and base types of MLOs involved.

\textbf{Baselines.} 
We compare \conveinsum-assisted training against two baselines across all tasks: \textbf{naive left-to-right} evaluation of tensorial forward passes with and without checkpointing.
In particular, we compare the usage of \conveinsum to optimally evaluate tensorial forward passes against these naive implementations to demonstrate the benefits of evaluating MLOs in a FLOPs-minimal order. For all backpropagation, autograd is used.

A compression rate (CR) of $x\%$ indicates that the size of each learnable layer of the TNN model $x\%$ the number of original layer. To achieve this, we first form the specified tensor decomposition of the learnable layer with rank $R$ such that it is equal in size to the original layer. We then trim off $x\%$\footnote{In the case of $100\%$ compression, the ranks of the reshaped tensors are selected so as to match the original backbone model size with no further reduction. The performance of such a TNN will not be equivalent to an un-tensorized counterpart despite containing the same number of parameters, since it is merely an approximation.} of the least significant components, i.e., reduce the rank, until it contains $\leq x\%$ of the original parameters.

We stress that this work is not focused on improving the accuracy of TNNs, which is an orthogonal topic of research. Accuracy drop, especially due to degeneracy in tensor decompositions, is a well-known phenomenon independent of MLO evaluation. In any case, we report TNN accuracy for various tasks as supplementary experimental data in Appendix \ref{app:experiments}.

%Many papers have improved TNN model accuracy using a wide variety of corrective mechanisms \citep{phan2020stable, su2018tensorial, jaderberg2014speeding, lee2021qttnet, goyal2019compression}.

%but for future work, we will investigate incorporating quantization, pruning, and knowledge distillation techniques directly into \autotnn.}

%As indicated in \Cref{tab:max-batch-size}, even an incorporation of checkpointing into \pytorch implementation does not help much with the problem.
%The reason for this is that although checkpointing sidesteps having to save all intermediate results during the forward pass, the order of evaluation in the forward pass remains unchanged. If an intermediary computation causes overflow, this issue is likely to persist even in a checkpointed implementation -- only the optimal sequencer can help with such a dilemma. 

%On average, the most cost-parsimonious path found by the optimal sequencer will contain smaller intermediate products than a naive left-to-right \pytorch evaluation. 

\subsection{Runtime and FLOPs results}
\label{sub:runtime}
We used \conveinsum and its optimal sequencer to evaluate a tensorial forward pass in an order which incurs the minimum number of FLOPs. A naive implementation will evaluate a tensorial sequence from left to right as given. Additionally, \conveinsum uses gradient checkpointing by default to avoid memory overflow for backward passes. 

\begin{figure}[!htbp]
\begin{minipage}[t]{0.45\linewidth}
\centering
\resizebox{\textwidth}{!}{
   \includegraphics[]{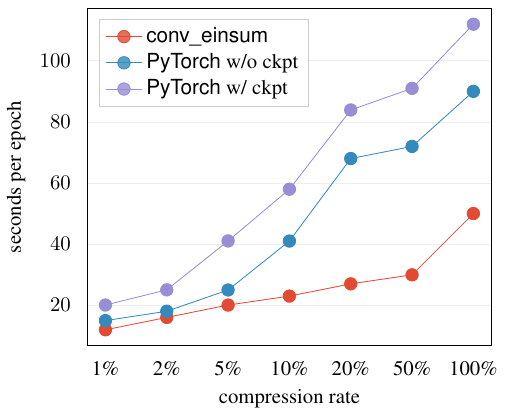}
    }
\subcaption{ASR Train}\label{fig:speech-rcp-vs-pytorch-train}
%\end{minipage}
%\hfill
%\begin{minipage}[t]{0.4\linewidth}
%\centering
\resizebox{\textwidth}{!}{
    \includegraphics[]{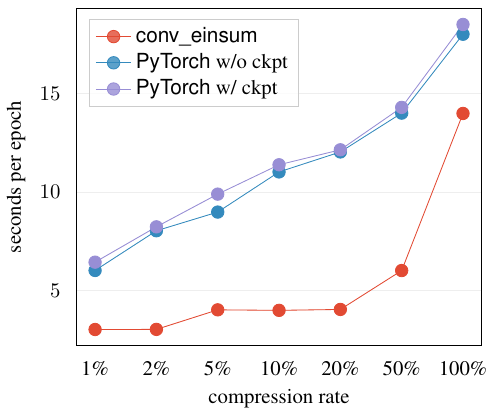}
    }
\subcaption{ASR Test}\label{fig:speech-rcp-vs-pytorch-test}
\end{minipage}
\hfill
\begin{minipage}[t]{0.45\linewidth}
\centering
\resizebox{\textwidth}{!}{
    \includegraphics[]{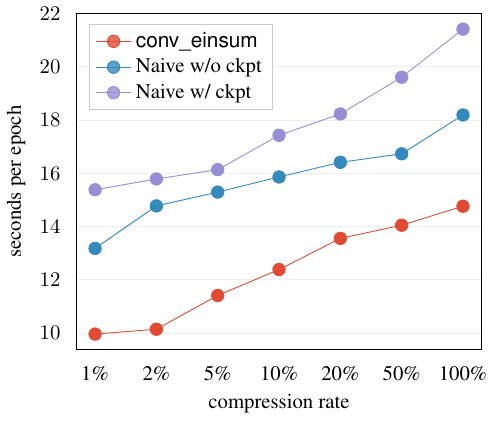}
    }
\subcaption{IC Train}
%\end{minipage}
%\hfill
%\begin{minipage}[t]{0.4\linewidth}
%\centering
\resizebox{\textwidth}{!}{
    \includegraphics[]{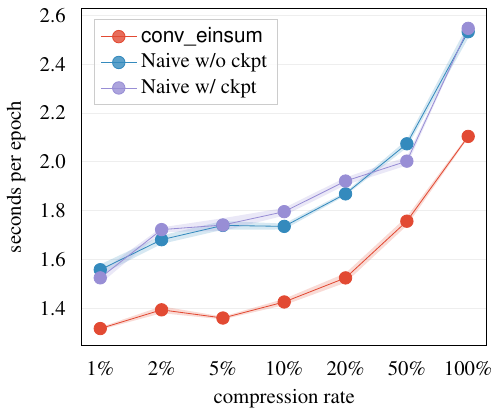}
    }
\subcaption{IC Test}
\end{minipage}
%\vspace{-2em}
\caption{\textbf{Run-time comparison between \conveinsum and naive evaluation for image classification (IC) and automatic speech recognition (ASR) (CP-TNN) tasks}. A RCP-TNN ($M=3$) and CIFAR-10 is used for IC while a CP-TNN and LibriSpeech dataset isused for ASR. 'ckpt' denotes checkpointing. Runtimes are averaged over 3 runs.}\label{fig:speech-rcp-vs-pytorch}
\end{figure}
\begin{figure}[t]
\begin{minipage}[t]{.45\linewidth}
\centering
\resizebox{\textwidth}{!}{
   \includegraphics[]{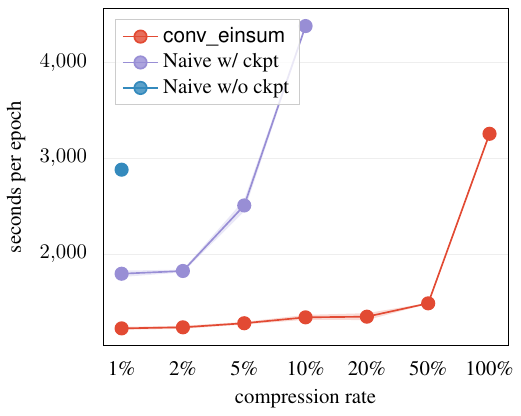}
    }
\subcaption{Train (spatial)}\label{fig:video-rcp-vs-pytorch-train-spatial}
%\end{minipage}%
\hfill
%\begin{minipage}[t]{.22\linewidth}
\resizebox{\textwidth}{!}{
    \includegraphics[]{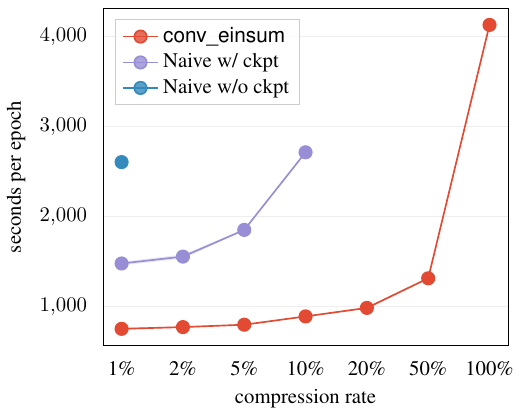}
    }
\subcaption{Test (spatial)}\label{fig:video-rcp-vs-pytorch-test-spatial}
\end{minipage}
\hfill
\begin{minipage}[t]{.45\linewidth}
\centering
\resizebox{\textwidth}{!}{
   \includegraphics[]{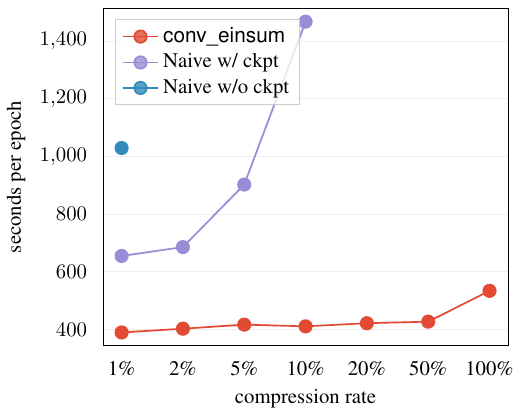}
    }
\subcaption{Train (temporal)}\label{fig:video-rcp-vs-pytorch-train-temporal}
%\end{minipage}%
\hfill
%\begin{minipage}[t]{.22\linewidth}
\resizebox{\textwidth}{!}{
    \includegraphics[]{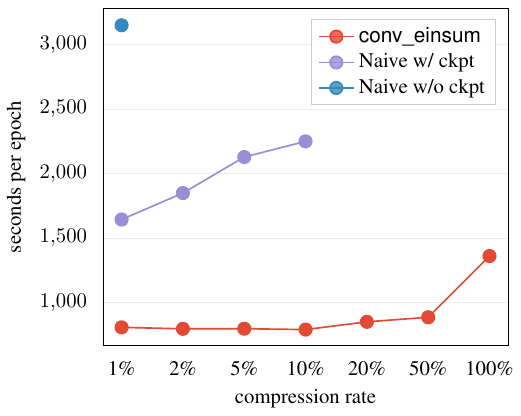}
    }
\subcaption{Test (temporal)}\label{fig:video-rcp-vs-pytorch-test-temporal}
\end{minipage}

\caption{\textbf{Run-time comparison between \conveinsum and naive left-to-right \pytorch implementations for a video classification machine learning task} An RCP-TNN ($M=3$) is trained on the UCF-101 dataset. All tests were run using the maximum allowable batch size. Naive w/ ckpt only ran without memory overflow for compression rates 1\% - 10\% and naive w/o ckpt was only able to run for compression rate 1\%. 'ckpt' denotes checkpointing. Averaged over 3 runs. }\label{fig:video-rcp-vs-pytorch}
\vspace{-1em}
\end{figure}
\begin{table}[!htbp]
\caption{\textbf{Run-time (minutes per epoch) comparison between \conveinsum and \pytorch for ImageNet Classiciation w/ checkpointing.} We use a RCP ($M=3$) ResNet-34 architecture and batch size of 256. Left-to-right evaluation runs into memory overflow without checkpointing. \conveinsum significantly improves training time. Runtimes are averaged over 3 runs.}
  \centering
\resizebox{0.5\linewidth}{!}{  \begin{tabular}{lcccc}
    \toprule
    & \multicolumn{2}{c}{\conveinsum} & \multicolumn{2}{c}{\shortstack{Naive w/ ckpt}} \\
    \cmidrule(lr){2-3}\cmidrule(lr){4-5}
    \thead{CR} & \thead{Train} & \thead{Test} & \thead{Train} & \thead{Test} \\
    \midrule
    5\% & 21.8 & 1.06 & 32.4 & 1.42 \\
    10\% & 23.4 & 1.47 & 35.7 & 1.77 \\
    20\% & 28.9 & 1.95 & 39.9 & 2.35 \\
    50\% & 34.1 & 2.54 & 50.1 & 3.54 \\
    100\% & 41.6 & 3.08 & 65.6 & 4.08 \\
    \bottomrule
\end{tabular}}
\label{tab:imagenet-rcp}
\end{table}

\emph{(1) The optimal sequencer used in \conveinsum significantly improves the runtime efficiency of training and test in TNNs.} 
In Figures \ref{fig:speech-rcp-vs-pytorch},\ref{fig:video-rcp-vs-pytorch} and Table \ref{tab:imagenet-rcp}, we compare the training and test times between \conveinsum and naive left-to-right (with and w/o checkpointing) implementations over a range of model scales and decompositions. The IC and VC tasks use RCP ($M=3)$ decompositions, while the ASR task uses a standard CP decomposition. 
In the VC task, for each model size, we use the maximal allowable batch size, while in the ASR task, we use the same batch size across scales. 
We observe that when memory is the bottleneck of a task (such as VC), checkpointing helps accelerate the runtime by allowing more batches. On the other hand, when the batch sizes are the same (as in ASR), checkpointing trades computational complexity for space, thus increasing the overall runtime. In either scenario, \conveinsum achieves the fastest runtimes.

\emph{(2) \conveinsum improves computation through weight tensors of differing sizes and decompositions.}
\conveinsum serves as a general solution for improving efficiency and is tensor-structure-agnostic. The networks we have experimented with contain weights of vastly differing sizes. As shown in \Cref{fig:speech-rcp-vs-pytorch,fig:video-rcp-vs-pytorch}, \conveinsum exhibits competitive results against standard left-to-right implementation (with or without) checkpointing over various tensor scales. Furthermore, \Cref{tab:different-tensor-forms,tab:runtime-cpu} in the Appendix shows the results of our \conveinsum training using different forms of tensor decomposition in both a high-performance (GPU) and low-performance (CPU) environment. We observe that \conveinsum outperforms naive implementation with/without checkpointing in all cases. 

\emph{(3) \conveinsum greatly reduces the number of FLOPs needed to process tensorial convolutional layers.}
Given a forward pass through a tensorial convolutional layer expressible as a \conveinsum string, an optimal evaluation path can be determined. Evaluation of an MLO sequence from left-to-right will result in an excessive amount of FLOPs as predicted by Theorem \ref{flop-reduce} and shown in Table \ref{tab:flop-costs}, which lists the number FLOPs saved by optimal path evaluation through CP convolutional layer blocks of ResNet-34.

\begin{table}[!htbp]
 \caption{\protect\textbf{FLOPs per CP convolutional layer in ResNet-34.} We calculate the FLOPs incurred by forward passes through CP convolutional layers (CR$=100\%$), using a batch size of 128. Layer nomenclature and associated kernel structure correspond with architecture description in \protect\cite{he2016deep}.} 
  \centering
\resizebox{0.6\linewidth}{!}{\begin{tabular}{llll}
    \toprule
    \thead{Layer} & \thead{Left-to-Right} & \thead{\conveinsum}  & \thead{Speedup}$\times$ \\
    \midrule
    conv1   & $1.51\times10^{15}$ & $3.87\times10^{13}$ & 3.90  \\
    conv2$\_$x    & $1.38\times10^{14}$ & $3.09\times10^{13}$  & 4.47  \\
    conv3$\_$x    & $5.85\times10^{13}$ & $9.67\times10^{12}$   & 6.05   \\
    conv4$\_$x    & $6.03\times10^{13}$  & $3.71\times10^{12}$   & 16.25   \\
    conv5$\_$x     & $4.25\times10^{13}$ & $4.72\times10^{11}$  &  90.04 \\
    \bottomrule
 \end{tabular}}

 \label{tab:flop-costs}
 
\end{table}

\subsection{Memory Results}
\label{sub:accuracy-memory}

% Figure for maximum batch size

\emph{Naive evaluation of tensorial forward passes suffers from high intermediary memory costs during training. \conveinsum significantly reduces these costs.}
In TNNs, during training computation of some intermediate data objects can be prohibitively large and thus difficult to fit into memory. In \Cref{tab:max-batch-size}, we present the maximal batch size allowed for two large-scale tasks under different compression rates: video classification (VC) and automatic speech recognition (ASR). We observe that if the size of a TNN matches the original backbone model size (i.e., CR=100\%), the maximal allowed batch size in a standard \pytorch implementation is 0 without checkpointing.
Even if we compress the model to $1\%$ of the original number of parameters, the maximal allowed batch size is still limited, making computation infeasible or slow. We theorize that optimal evaluation order results in uniformly smaller intermediate products.

\begin{table}[!tb]
\caption{\textbf{Maximum batch size for a speech and video task.} The maximal batch size allowed for data under varying compression rates and using different libraries on \textbf{(1)} an ASR on LibriSpeech and \textbf{(2)} a VC task for spatial (S) and temporal (T) streams of a two-stream network on UCF-101. \conveinsum permits larger batch sizes.  %using \texttt{conv\_einsum} and \pytorch with/without checkpointing.
}
 \label{tab:max-batch-size}
  \centering
  %\resizebox{\linewidth}{!}
  {
  \begin{tabular}[!tb]{lccc}
    %\toprule
    \multicolumn{4}{c}{\textbf{Automatic speech recognition task}}\\
    \hline
    \thead{CR} & \thead{\conveinsum} & \thead{{\shortstack{Naive \\ w/ ckpt}}} & \thead{{\shortstack{Naive \\w/o ckpt}} }\\
    \midrule
    1\% & 14 & 8 & 6\\
    2\% & 14 & 8 & 6\\
    5\% & 12 & 6 & 4\\
    10\% & 10 & 4 & 2\\
    20\% & 8 & 2 & 0\\
    50\% & 6 & 2 & 0 \\
    100\% & 4 & 1 & 0\\
  \bottomrule
 \end{tabular}}
 \label{tab:video-max-batch-size}
  \centering
  %\resizebox{\linewidth}{!}
  {
  \begin{tabular}[!tb]{lcccccc}
    %\toprule
     &&\multicolumn{3}{c}{\textbf{Video classification task}}
     \\
     \hline
     &\multicolumn{2}{c}{\conveinsum} & \multicolumn{2}{c}{\shortstack{Naive \\w/ ckpt}} & \multicolumn{2}{c}{\shortstack{Naive \\w/o ckpt}} \\
    \cmidrule(lr){2-3}\cmidrule(lr){4-5}\cmidrule(lr){6-7}
    \thead{CR} & \thead{S} & \thead{T} & \thead{S} & \thead{T} & \thead{S} & \thead{T}\\
    \midrule
    1\% & 20 & 30 & 2 & 4 & 1 & 2\\
    2\% & 20 & 30 & 2 & 4 & 0 & 0\\
    5\% & 20 & 30 & 2 & 4 & 0 & 0\\
    10\% & 18 & 30 & 1 & 2 & 0 & 0\\
    20\% & 16 & 27 & 0 & 0 & 0 & 0\\
    50\% & 12 & 22 & 0 & 0 & 0 & 0\\
    100\% & 4 & 14 & 0 & 0 & 0 & 0\\
    \bottomrule
 \end{tabular}}
\end{table}

%Maximum batch size for a video classification task for spatial (S) and temporal (T) streams of two-stream network.

\section{Conclusions and Discussions}
\label{sec:conclusion}

In this work, we introduce an \einsum-like notational framework and meta-algorithm \conveinsum, which is capable of expressing and efficiently computing passes through convolutional TNNs. \conveinsum-assisted training is competitive against, and in many cases, superior to standard \pytorch TNN training.
For future work, we plan to accelerate training and test times through incorporation of parallel computation libraries such as TensorRT. Additionally, our experiments do not make use of low-rank decomposition stabilizing algorithms \cite{phan2020stable,jaderberg2014speeding} which improve the quality of factorized components. Combining \conveinsum with such methods could result in exceptionally light, accurate, and fast TNNs.

\section*{Acknowledgments}
All authors, at the time of writing, acknowledge support by the National Science Foundation NSF-IIS-FAI program, DOD-ONR-Office of Naval Research, DOD-DARPA-Defense Advanced Research Projects Agency Guaranteeing AI Robustness against Deception (GARD), Adobe, Capital One and JP Morgan faculty fellowships. 
Rabbani was additionally supported by NSF DGE-1632976.
\newpage

\bibliographystyle{unsrt}
\bibliography{0_main.bib}
%\clearpage

%%%%%%%%%%%%%%%%%%%%%%%%%%%%%%%%%%%%%%%%%%%%%%%%%%%%%%%%%%%%

\clearpage
\appendix

% \begin{center}
% {\bf \Large Appendices for \title}
% \end{center}

%\section{Additional Related Work}\label{app:add_related_work}

\section*{Appendix}
\section{Supplementary Material for Tensor Operations and \einsum}
\label{app:A1multi}

\subsection{Examples of primitive operations}\label{app:subsec:einsum}

% 1) Tensor contraction
\textbf{Contraction.}
{\em Tensor contraction} generalizes matrix multiplication to higher-order tensors. 
For instance, given two $3^\rd$ order tensors $\tensorSup{T}{1} \in \R^{A \times B \times C}, \tensorSup{T}{2} \in \R^{A \times D \times E}$, we can define a contraction between the modes with shared dimension size $A$. The operation returns a $4^\th$ order tensor $\tensor{T} \in \R^{B \times C \times D \times E}$ with its entries calculated as:
%\begin{equation}
%\label{eq:contraction}
$\tensorSub{T}{b,c,d,e} = \sum_{a = 1}^{A}
\tensorInd{T}{1}{a,b,c} \cdot \tensorInd{T}{2}{a,d,e}$.
%\end{equation}
This contraction would be submitted to \einsum as:
\begin{lstlisting}
T = einsum("abc,ade->bcde", T1, T2)
\end{lstlisting}

Here, the modes are ordered and represented by concatenated letters (which can be case-sensitive). These sub-strings, corresponding to each input tensor, are separated by commas and lie to the left of the arrow. The output, which lies to the right of the arrow, concatenates all mode letters in a single string sans the letter $\textsf{"a"}$ to indicate that a contraction must occur over these modes. The remaining parameters correspond to the ordered set of tensors. This method will fail if the modes denoted by the same letters do not share dimension sizes. So a contraction corresponds to a letter that appears in all inputs, but not the output.

% 2) Tensor outer product
\textbf{Outer product.}
{\em Tensor outer product} generalizes outer product to higher-order tensors. For instance, an outer product of two $3^\rd$ order tensors $\tensorSup{T}{1} \in \R^{A \times B \times C}, \tensorSup{T}{2} \in \R^{D \times E \times F}$ returns a $6^\th$ order tensor $\tensor{T} \in \R^{A \times B \times C \times D \times E \times F}$. The entries of $\tensor{T}$ are calculated as:
%\begin{equation}
%\label{eq:outer-product}
$\tensorSub{T}{a,b,c,d,e,f} =
\tensorInd{T}{1}{a,b,c} \cdot
\tensorInd{T}{2}{d,e,f}$.
%\end{equation}
In the language of \einsum, we write this as:
\begin{lstlisting}
T = einsum("abc,def->abcdef", T1, T2)
\end{lstlisting}
\vspace{-1em}
Notice that a letter for the outer product only appears in one of the inputs and in the output.

% 3) Tensor batch product
\textbf{Batch product.}
{\em Tensor batch product} is a variation of the outer product. Given tensors $\tensorSup{T}{1} \in \R^{A \times B \times C}, \tensorSup{T}{2} \in \R^{C \times E \times D}$, the batch product over the dimensional-shared mode $C$ returns a $5^\th$ order tensor $\tensor{T}\in\R^{A \times B \times C \times D \times E}$. The entries of $\tensor{T}$ are calculated as:
%\begin{equation}
%\label{eq:batch-product}
$\tensorSub{T}{a,b,c,d,e} =
\tensorInd{T}{1}{a,b,c} \cdot
\tensorInd{T}{2}{a,d,e}$.
%\end{equation}
The \einsum implementation is
\begin{lstlisting}
T = einsum("abc,cde->abcde", T1, T2)
\end{lstlisting}
\vspace{-1em}
Notice that the letter corresponding to the batch product appears in both inputs and the output.

In this section, we will outline the multi-linear operations covered in preliminaries in full generality along with their \conveinsum representations and other tensor decompositions of TNNs. Much of this content adapts and follows the notation of Su et al.~\cite{su2018tensorial}.

\subsection{Fully General Multilinear operations} 
\label{app-sub:multiops}

% Multi-operations among multiple tensors
\textbf{Multi-operations among multiple tensors.}
We can simultaneously perform a series of multi-linear operations among a group of tensors.
Let $\tensorSup{T}{1} \in \R^{I \times T \times S}$,
$\tensorSup{T}{2} \in \R^{J \times R \times T}$, and
$\tensorSup{T}{3} \in \R^{K \times S \times R}$. We can define a simultaneous contraction on modes with dimension sizes $R, S,$ and $T$.
This simultaneous contraction returns a $3^\rd$ order tensor $\tensor{T} \in \R^{I \times J \times K}$, with its entries computed as
\begin{equation}
\label{eq:multi-operation-2}
\tensorSub{T}{i,j,k} = \sum_{r = 1}^{R} \sum_{s = 1}^{S} \sum_{t = 1}^{T} 
\tensorInd{T}{1}{i,t,s} \tensorInd{T}{2}{j,r,t} \tensorInd{T}{3}{k,s,r}
\end{equation}
Equivalently, via \conveinsum, we can write it as:
\begin{lstlisting}
T = conv_einsum("its,jrt,ksr->ijk", T1, T2, T3)
\end{lstlisting}
\vspace{-1em}
% In Appendix~\ref{multiops}, 
Below we outline a simpler example that performs multiple operations between two tensors.

% Table for primitive operations
\begin{table*}[!htbp]
\centering
\resizebox{\textwidth}{!}{
\setlength{\tabcolsep}{3pt}
\begin{tabular}{c| c |c}
\toprule
Operation & Definition & \conveinsum \\
\midrule
	% 1) tensor contraction
	\begin{tabular}{l} mode-($k, l$) \\ Tensor Contraction \end{tabular} & 
	\begin{tabular}{l} $\tensor{T}_{i_0, \cdots, i_{k-1}, i_{k+1}, \cdots, i_{m-1}, j_0, \cdots, j_{l-1}, j_{l+1}, \cdots, j_{n-1}}$ \\
	$ = \big< \tensorSup{T}{0}_{i_0, \cdots, i_{k-1},{ :}, i_{k+1}, \cdots, i_{m-1}},  \tensorSup{T}{1}_{j_0, \cdots, j_{l-1}, {:}, j_{l+1}, \cdots, j_{n-1}}\big>$ \\ 
	\end{tabular} & \begin{tabular}{l} $\textcolor{red}{I_0 \cdots I_{k-1} I_k \cdots I_{m-1}}, \textcolor{blue}{J_0 \cdots J_{l-1} J_l \cdots J_{n-1}}$ \\ $\rightarrow  I_{1} I_{k-1} I_{k+1} I_{m-1} J_0 \cdots J_{l-1} J_{l+1} \cdots J_{n-1}$ \end{tabular}\\
\midrule
	% 2) tensor convolution
	\begin{tabular}{l} mode-$(k, l)$ \\Tensor Convolution \end{tabular} & 
	\begin{tabular}{l}$\tensorInd{T}{i_0, \cdots, i_{k-1}, {:}, i_{k+1}, \cdots, i_{m-1}, j_0, \cdots, j_{l-1}, j_{l+1}, \cdots, j_{n-1}}{2} $ \\
	$= \tensorSup{T}{0}_{i_0, \cdots, i_{k-1}, {:}, i_{k+1}, \cdots, i_{m-1}} \ast \tensorSup{T}{1}_{j_0, \cdots, j_{l-1},{ :}, j_{l+1}, \cdots, j_{n-1}}$ \\ 
	 \end{tabular} & \begin{tabular}{l} $\textcolor{red}{I_0 \cdots I_{k-1} C \cdots I_{m-1}}, \textcolor{blue}{J_0 \cdots J_{l-1} C \cdots J_{n-1}}$ \\ $\rightarrow  I_{1} I_{k-1} C I_{k+1} I_{m-1} J_0 \cdots J_{l-1} J_{l+1} \cdots J_{n-1} \mid C$ \end{tabular}\\
\midrule
	% 3) tensor batch product
	\begin{tabular}{l} mode-($k, l$) \\ Tensor Batch Product \end{tabular} &
	\begin{tabular}{l} $\tensor{T}_{i_0, \cdots, i_{k-1}, {r}, i_{k+1}, \cdots, i_{m-1}, j_0, \cdots,  j_{n-1}} $ \\
	$=\tensorSup{T}{0}_{i_0, \cdots, i_{k-1}, {r}, i_{k+1}, \cdots, i_{m-1}} ~ \tensorSup{T}{1}_{j_0, \cdots, j_{l-1},{ r}, j_{l+1}, \cdots, j_{n-1}}$ \\ 
	\end{tabular} & \begin{tabular}{l} $\textcolor{red}{I_0 \cdots I_{k-1} C \cdots I_{m-1}}, \textcolor{blue}{J_0 \cdots J_{l-1} J_l \cdots J_{n-1}}$ \\ $\rightarrow  I_{1} I_{k-1} I_k I_{k+1} I_{m-1} J_0 \cdots J_{l-1} J_{l+1} \cdots J_{n-1}$  \end{tabular}\\
\midrule
	% 2) tensor outer product
	\begin{tabular}{l} Tensor Outer Product \end{tabular} & 
	\begin{tabular}{l}  $\tensorSup{T}{0}_{i_0, \cdots, i_{k-1}, i_k, i_{k+1}, \cdots, i_{m-1}, j_0, \cdots, j_{l-1}, j_l, j_{l+1}, \cdots, j_{n-1}}$ \\
	$= \tensorSup{T}{1}_{i_0, \cdots, i_{m-1}} \ast \tensorSub{Y}{j_0, \cdots, j_{l+1}, \cdots, j_{n-1}}$ \\ 
	 \end{tabular} &
	 \begin{tabular}{l} $\textcolor{red}{I_0 \cdots I_{m-1}}, \textcolor{blue}{J_0 \cdots J_{n-1}}$ \\ $\rightarrow  I_{1} \cdots I_{m-1} J_0 \cdots J_{n-1}$  \end{tabular}\\
\bottomrule
\end{tabular}
}
\caption{\textbf{Primitive tensor operations}. 
$\tensorSup{T}{0} \in \R^{I_0 \times \cdots \times I_{m-1}}, 
\tensorSup{T}{1} \in \R^{J_0 \times \cdots \times J_{n-1}}$ are the input tensors and
$\tensor{T}$ is the output tensor.
Note that both mode-$(I_k, J_l)$ tensor contraction, mode-($I_k, J_l$) tensor batch product 
are legal only if $I_k = J_l$. The outer product can be performed with any two tensors. Red highlighted modes are owned by $\tensorSup{T}{0}$ while blue highlighted modes belong to $\tensorSup{T}{1}$.}
\label{tab:primitive-operations}
\end{table*}

% (1) Tensor contraction
\paragraph{Tensor contraction} 
Given tensors $\tensorSup{T}{0}\in\mathbb{R}^{I_0 \times \cdots I_k \times I_{m-1}}, \tensorSup{T}{1}\in\mathbb{R}^{J_0 \times \cdots J_l  \cdots \times J_{n-1}}$, the mode-($k,l$) contraction returns an order $(m+n-2)$ tensor \\$\tensor{T}\in\mathbb{R}^{I_0\times\cdots I_{k-1} \times I_{k+1}\times \cdots \times I_{m-1}\times J_0 \times \cdots \times J_{l-1}\times J_{l+1}\times\cdots\times J_{n-1} }$. The entries of $\tensor{T}$ are calculated as follows: 
\begin{align*}
&\tensor{T}_{i_0, \cdots, i_{k-1}, i_{k+1}, \cdots, i_{m-1}, j_0, \cdots, j_{l-1}, j_{l+1}, \cdots, j_{n-1}}\\ &=\sum_{r=1}^{I_k-1}\tensorSup{T}{0}_{i_0, \cdots, i_{k-1},r, i_{k+1}, \cdots, i_{m-1}}\cdot  \tensorSup{T}{1}_{j_0, \cdots, j_{l-1}, r, j_{l+1}, \cdots, j_{n-1}}\\
&=\langle \tensorSup{T}{0}_{i_0, \cdots, i_{k-1},{ :}, i_{k+1}, \cdots, i_{m-1}},  \tensorSup{T}{1}_{j_0, \cdots, j_{l-1}, {:}, j_{l+1}, \cdots, j_{n-1}}\rangle.
\end{align*}

% (2) Tensor convolution
\paragraph{Tensor Convolution} 
Given tensors $\tensorSup{T}{0}\in\mathbb{R}^{I_0\times \cdots \times I_{m-1}}, \tensorSup{T}{1}\in\mathbb{R}^{J_0 \times \cdots \times J_{n-1}}$, the mode-($k,l$) convolution returns an order $(m+n-1)$ tensor \\$\tensor{T}\in\mathbb{R}^{I_0\times\cdots I'_k \times \cdots \times I_{m-1}\times J_0 \times \cdots \times J_{l-1}\times J_{l+1}\times\cdots\times J_{n-1}}$. For any convolution operator * the entries of $\tensor{T}$ are calculated as follows:
\begin{align*}
& \tensor{T}_{i_0, \cdots, i_{k-1}, {:}, i_{k+1}, \cdots, i_{m-1}, j_0, \cdots, j_{l-1}, j_{l+1}, \cdots, j_{n-1}} \\
	&= \tensorSup{T}{0}_{i_0, \cdots, i_{k-1}, {:}, i_{k+1}, \cdots, i_{m-1}} \ast \tensorSup{T}{1}_{j_0, \cdots, j_{l-1},{ :}, j_{l+1}, \cdots, j_{n-1}}\\
	&= \tensorSup{T}{1}_{j_0, \cdots, j_{l-1},{ :}, j_{l+1}, \cdots, j_{n-1}} \bar{\ast} \tensorSup{T}{0}_{i_0, \cdots, i_{k-1}, {:}, i_{k+1}, \cdots, i_{m-1}}
\end{align*}
Here we have intentionally left $\ast$ and the dimension of the $k$-th mode $I'_k$ ambiguous, as it will vary depending on the type of convolution specified by the user. For example, with max padding, we have that $I'_k=\max\{I_k,J_l\}$, and with same-padding we have $I'_k=I_k$. 

% (3) tensor batch product
\paragraph{Tensor Batch Product}
Given tensors $\tensorSup{T}{0}\in\mathbb{R}^{I_0\times \cdots I_{k} \cdots \times I_{m-1}}, \tensorSup{T}{1}\in\mathbb{R}^{J_0 \times \cdots I_l \cdots \times J_{n-1}}$, the mode-($k,l$) batch product returns an order $(m+n-1)$ tensor $\tensor{T}\in\mathbb{R}^{I_0\times\cdots I_k \times \cdots \times I_{m-1}\times J_0 \times \cdots \times \cdots\times J_{n-1}}$. The entries of $\tensor{T}$ are calculated as follows:
\begin{align*}
& \tensor{T}_{i_0, \cdots, i_{k-1}, r, i_{k+1}, \cdots, i_{m-1}, j_0, \cdots, j_{l-1}, j_{l+1}, \cdots, j_{n-1}} \\
	&= \tensorSup{T}{0}_{i_0, \cdots, i_{k-1}, r, i_{k+1}, \cdots, i_{m-1}} \tensorSup{T}{1}_{j_0, \cdots, j_{l-1},r, j_{l+1}, \cdots, j_{n-1}}
\end{align*}

% (4) tensor outer product
\paragraph{Tensor Outer Product}
 Given tensors $\tensorSup{T}{0}\in\mathbb{R}^{I_0\times \cdots \times I_{m-1}}, \tensorSup{T}{1}\in\mathbb{R}^{J_0 \times \cdots \times J_{n-1}}$, the outer product returns an order $(m+n)$ tensor $\tensor{T}\in\mathbb{R}^{I_0\times\cdots I_k \times \cdots \times I_{m-1}\times J_0 \times \cdots \times J_{n-1}}$. The entries of $\tensor{T}$ are calculated as follows:
\begin{align}
\tensor{T}_{i_0, \cdots, i_{m-1}, j_0, \cdots, j_{n-1}}
	&= \tensorSup{T}{0}_{i_0, \cdots, i_{m-1}} \tensorSup{T}{1}_{j_0, \cdots, j_{n-1}}.
\end{align}

\Cref{tab:primitive-operations} summarizes these multi-linear operations along with their \conveinsum input representations. 

\subsection{Tensorial Neural Networks}
\label{app-sub:TNN}

% 1) CP decomposition
\textbf{CP decomposition}~\cite{kolda2009tensor}.
(1a) In a {\em CP convolutional layer}~\cite{lebedev2015speeding}, the original kernel $\tensor{W} \in \R^{T \times S \times H \times W}$ is CP factorized as:
\begin{lstlisting}
W = conv_einsum("rt,rs,rh,rw->tshw", W1, W2, W3, W4).
\end{lstlisting}
\vspace{-0.5em}
As a result, the layer is parameterized by $4$ weight matrices $\matrixSup{W}{1} \in \R^{R \times T}$, $\matrixSup{W}{2} \in \R^{R \times S}$, $\matrixSup{W}{3} \in \R^{R \times H}$, $\matrixSup{W}{4} \in \R^{R \times W}$. We can write this layer in \conveinsum as
\begin{lstlisting}
Y = conv_einsum("bshw,rt,ts,rh,rw->bthw|hw", X, W1)
\end{lstlisting}
\vspace{-0.5em}

(1b) In a {\em reshaped CP convolutional layer}~\cite{su2018tensorial}, the original kernel $\mytensor{W}$ is first reshaped as $\mytensor{\overline{W}} \in \R^{T_1 \cdots \times T_M \times S_1 \cdots \times S_M \times H \times W}$ such that $T = \prod_{m = 1}^{M} T_m$ and $S = \prod_{m = 1}^{M} S_m$. Suppose $M = 3$, then the reshaped kernel is factorized by a CP decomposition as
\begin{lstlisting}
W = conv_einsum("r(t1)(s1),r(t2)(s2),r(t3)(s3),rhw->(t1)(t2)(t3)(s1)(s2)(s3)hw", W1, W2, W3, W0).
\end{lstlisting}
\vspace{-0.5em}
As a result, the layer is parameterized by $(M + 1)$ weight tensors $\tensorSup{W}{m} \in \R^{R \times T_m \times S_m}$ and $\tensorSup{W}{0} \in \R^{R \times H \times W}$. We can write the layer in \conveinsum as
\begin{lstlisting}
Y = conv_einsum("b(s1)(s2)(s3)hw,r(t1)(s1),r(t2)(s2),r(t3)(s3),rhw->b(t1)(t2)(t3)hw|hw", X, W1, W2, W3, W0)
\end{lstlisting}

% 2) Tucker decomposition 
\textbf{Tucker (TK) decomposition}~\cite{kolda2009tensor}.
(2a) In a {\em TK convolutional layer}~\cite{lebedev2015speeding}, the original kernel $\tensor{W} \in \R^{T \times S \times H \times W}$ is factorized by TK as:
\begin{lstlisting}
W = conv_einsum("(r1)t,(r2)s,(r1)(r2)hw->tshw", W1, W2, W0)
\end{lstlisting}
\vspace{-0.5em}
Consequently, the layer has $3$ weight tensors $\matrixSup{W}{1} \in \R^{R_1 \times T}$, $\matrixSup{W}{2} \in \R^{R_2 \times S}$, $\tensorSup{W}{0} \in \R^{R_1 \times R_2 \times H \times W}$ as parameters. We can write the layer in \conveinsum as
\begin{lstlisting}
Y = conv_einsum("bshw,(r1)t,(r2)s,(r1)(r2)hw->bthw|hw", X, W1, W2, W0)
\end{lstlisting}
\vspace{-0.5em}

(2b) In a {\em reshaped TK convolutional layer}~\cite{su2018tensorial}, the original kernel $\mytensor{W}$ is first reshaped as $\mytensor{\overline{W}} \in \R^{T_1 \cdots \times T_M \times S_1 \cdots \times S_M \times H \times W}$ such that $T = \prod_{m = 1}^{M} T_m$ and $S = \prod_{m = 1}^{M} S_m$. Suppose $M = 3$, the reshaped kernel is then factorized by a TK decomposition as
\begin{lstlisting}
W = conv_einsum("(r1)(t1)(s1),(r2)(t2)(s2),(r3)(t3)(s3),(r0)hw,(r0)(r1)(r2)(r3)->(t1)(t2)(t3)(s1)(s2)(s3)hw", W1, W2, W3, W0, C)
\end{lstlisting}
\vspace{-0.5em}
Therefore, the layer has $(M + 2)$ weight tensors $\tensorSup{W}{m} \in \R^{R \times T_m \times S_m}$, $\tensorSup{W}{0} \in \R^{R \times H \times W}$, $\tensor{C} \in \R^{R_0 \times R_1 \times R_2 \times R_3}$. We can write the layer in \conveinsum as
\begin{lstlisting}
Y = conv_einsum("b(s1)(s2)(s3)hw,(r1)(t1)(s1),(r2)(t2)(s2),(r3)(t3)(s3),(r0)hw,(r0)(r1)(r2)(r3)->b(t1)(t2)(t3)hw|hw", X, W1, W2, W3, W0, C)
\end{lstlisting}

% Tensor-Train decomposition
\textbf{Tensor-Train (TT) decomposition}~\cite{oseledets2011tensor}.
(3a) In a {\em TT convolutional layer}, the original kernel $\tensor{W} \in \R^{T \times S \times H \times W}$ is factorized by TT as:
\begin{lstlisting}
W = conv_einsum("(r1)t,(r1)(r2)h,(r2)(r3)w,(r3)s->tshw", W1, W2, W3, W4)
\end{lstlisting}
\vspace{-1em}
Consequently, the layer has $4$ weight tensors $\matrixSup{W}{1} \in \R^{R_1 \times T}$, $\matrixSup{W}{2} \in \R^{R_1 \times R_2 \times H}$, $\tensorSup{W}{3} \in \R^{R_2 \times R_3 \times W}$, and $\tensorSup{W}{4} \in \R^{R_3 \times S}$ as parameters. We can write the layer in \conveinsum as
\begin{lstlisting}
Y = conv_einsum("bshw,(r1)t,(r1)(r2)h,(r2)(r3)w,(r3)s->bthw|hw", X, W1, W2, W3, W4).
\end{lstlisting}
\vspace{-0.5em}

(3b) In a {\em reshaped TT convolutional layer}~\cite{garipov2016ultimate}, the original kernel $\mytensor{W}$ is first reshaped as $\mytensor{\overline{W}} \in \R^{T_1 \cdots \times T_M \times S_1 \cdots \times S_M \times H \times W}$ such that $T = \prod_{m = 1}^{M} T_m$ and $S = \prod_{m = 1}^{M} S_m$. Suppose $M = 3$, the reshaped kernel is then factorized by a TT decomposition as
\begin{lstlisting}
W = conv_einsum("(r1)(t1)(s1),(r1)(r2)(t2)(s2),(r2)(r3)(t3)(s3),(r3)hw->(t1)(t2)(t3)(s1)(s2)(s3)hw", W1, W2, W3, W0).
\end{lstlisting}

Therefore, the layer has $(M + 1)$ weight tensors $\tensorSup{W}{1} \in \R^{R_1 \times T_1 \times S_1}$, $\tensorSup{W}{m} \in \R^{R_{m - 1} \times R_m \times T_m \times S_m}$, and $\tensorSup{W}{0} \in \R^{R_3 \times H \times W}$. The layer in \conveinsum is
\begin{lstlisting}
Y = conv_einsum("b(s1)(s2)(s3)hw,(r1)(t1)(s1),(r1)(r2)(t2)(s2),(r2)(r3)(t3)(s3),(r3)hw->b(t1)(t2)(t3)hw|hw", X, W1, W2, W3, W0).
\end{lstlisting}

% Tensor-Ring decomposition
\textbf{Tensor-Ring (TR) decomposition}~\cite{zhao2016tensor}.
(4a) In a {\em TR convolutional layer}, the original kernel $\tensor{W} \in \R^{T \times S \times H \times W}$ is factorized by TR as:
\begin{lstlisting}
W = conv_einsum("(r0)(r1)t,(r1)(r2)h,(r2)(r3)w,(r3)(r0)s->tshw", W1, W2, W3, W4)
\end{lstlisting}

Consequently, the layer has $4$ weight tensors $\matrixSup{W}{1} \in \R^{R_1 \times T}$, $\matrixSup{W}{2} \in \R^{R_0 \times R_1 \times R_2 \times H}$, $\tensorSup{W}{3} \in \R^{R_2 \times R_3 \times W}$, and $\tensorSup{W}{4} \in \R^{R_3 \times R_0 \times S}$ as parameters. We can write the layer in \conveinsum as
\begin{lstlisting}
Y = conv_einsum("bshw,(r0)(r1)t,(r1)(r2)h,(r2)(r3)w,(r3)(r0)s->bthw|hw", X, W1, W2, W3, W4)
\end{lstlisting}

(4b) In a {\em reshaped TR convolutional layer}~\cite{su2018tensorial}, the original kernel $\mytensor{W}$ is first reshaped as $\mytensor{\overline{W}} \in \R^{T_1 \cdots \times T_M \times S_1 \cdots \times S_M \times H \times W}$ such that $T = \prod_{m = 1}^{M} T_m$ and $S = \prod_{m = 1}^{M} S_m$. Suppose $M = 3$, the reshaped kernel is then factorized by a TR decomposition as
\begin{lstlisting}
W = conv_einsum("(r0)(r1)(t1)(s1),(r1)(r2)(t2)(s2),(r2)(r3)(t3)(s3),(r3)(r0)hw->(t1)(t2)(t3)(s1)(s2)(s3)hw", W1, W2, W3, W0)
\end{lstlisting}
\vspace{-0.5em}
Therefore, the layer has $(M + 1)$ weight tensors $\tensorSup{W}{m} \in \R^{R_{m - 1} \times R_m \times T_m \times S_m}$, and $\tensorSup{W}{0} \in \R^{R_3 \times R_0 \times H \times W}$. The layer in \conveinsum is
\begin{lstlisting}
Y = conv_einsum("b(s1)(s2)(s3)hw,(r0)(r1)(t1)(s1),(r1)(r2)(t2)(s2),(r2)(r3)(t3)(s3),(r3)(r0)hw->b(t1)(t2)(t3)hw|hw", X, W1, W2, W3, W0)
\end{lstlisting}

\textbf{Block-Term (BT) decomposition}~\cite{ye2020block}. In a {\em reshaped BT convolutional layer}, $\mytensor{\overline{W}} \in \R^{T_1 \cdots \times T_M \times S_1 \cdots \times S_M \times H \times W}$ such that $T = \prod_{m = 1}^{M} T_m$ and $S = \prod_{m = 1}^{M} S_m$. Suppose $M = 3$, the reshaped kernel is then factorized by a BT decomposition as
\begin{lstlisting}
\textcolor{blue}{W = conv_einsum("r(r1)(t1)(s1),r(r2)(t2)(s2),r(r3)(t3)(s3),r(r0)hw,r(r1)(r2)(r3)(r0)->(t1)(t2)(t3)(s1)(s2)(s3)hw", W1, W2, W3, W0, C)}
\end{lstlisting}
\vspace{-0.5em}
Consequently, the layer has $(m+1)$ weight tensors $\tensorSup{W}{1} \in \R^{R \times R_1 \times T_1 \times S_1}$, $\tensorSup{W}{2} \in \R^{R \times R_2 \times T_2 \times S_2}$, $\tensorSup{W}{3} \in \R^{R \times R_3 \times T_3 \times S_3}$, $\tensorSup{W}{0} \in \R^{R \times R_0 \times H \times W}$, and $\mytensor{C} \in \R^{R \times R_1 \times R_2 \times R_3 \times R_0}$. The layer in \conveinsum is
\begin{lstlisting}
Y = conv_einsum("b(s1)(s2)(s3)hw,r(r1)(t1)(s1),r(r2)(t2)(s2),r(r3)(t3)(s3),r(r0)hw,r(r1)(r2)(r3)(r0)->b(t1)(t2)(t3)hw|hw", X, W1, W2, W3, W0, C)
\end{lstlisting}

\textbf{Hierarchical-Tucker (HT) decomposition}~\cite{wu2020hybrid}. In a {\em reshaped HT convolutional layer}, $\mytensor{\overline{W}} \in \R^{T_1 \cdots \times T_M \times S_1 \cdots \times S_M \times H \times W}$ such that $T = \prod_{m = 1}^{M} T_m$ and $S = \prod_{m = 1}^{M} S_m$. Suppose $M = 3$, the reshaped kernel is then factorized by a HT decomposition as
\begin{lstlisting}
W = conv_einsum("(r1)(t1)(s1),(r2)(t2)(s2),(r3)(t3)(s3),(r0)hw,(r1)(r2)(r4),(r3)(r0)(r5),(r4)(r5)->(t1)(t2)(t3)(s1)(s2)(s3)hw", W1, W2, W3, W0, C1, C2, C3)
\end{lstlisting}
\vspace{-0.5em}
Consequently, the layer has $(2m+1)$ weight tensors $\tensorSup{W}{1} \in \R^{R_1 \times T_1 \times S_1}$, $\tensorSup{W}{2} \in \R^{R_2 \times T_2 \times S_2}$, $\tensorSup{W}{3} \in \R^{R_3 \times T_3 \times S_3}$, $\tensorSup{W}{0} \in \R^{R_0 \times H \times W}$, $\tensorSup{C}{1} \in \R^{R_1 \times R_2 \times R_4}$, $\tensorSup{C}{2} \in \R^{R_3 \times R_0 \times R_5}$, and $\matrixSup{C}{3} \in \R^{R_4 \times R_5}$. 
The layer in \conveinsum is
\begin{lstlisting}
Y = conv_einsum("b(s1)(s2)(s3)hw,(r1)(t1)(s1),(r2)(t2)(s2),(r3)(t3)(s3),(r0)hw,(r1)(r2)(r4),(r3)(r0)(r5),(r4)(r5)->b(t1)(t2)(t3)hw|hw", X, W1, W2, W3, W0, C1, C2, C3)
\end{lstlisting}

\subsubsection{Efficient Convolutional Layers} \label{app:efficientNN}
A number of works design efficient convolutional layers by modifying the linear operations in deep networks. These types of convolutional layers can be thought as special cases of TNNs. 
Our proposed \conveinsum can cover these alternative efficient designs as well. 
Here, we review two representative designs, namely the {\em interleaved group convolution}~\cite{zhang2017interleaved} and {\em separable depth-wise convolution}~\cite{chollet2017xception}, in the language of \conveinsum.

% group interleaved convolution
{\bf (1)} In an {\em interleaved group convolution}, the layer partitions the input channels $S$ into two modes $M$ and $S^\prime$ such that $S = M S^\prime$, and the output channel $T$ into two modes $N$ and  $T^\prime$ such that $T = N T^\prime$.
As a result, the layer has two $4^\th$ order tensors $\tensorSup{W}{1} \in \R^{N \times M \times H \times W}$, $\tensorSup{W}{2} \in \R^{T^\prime \times S^\prime \times H \times W}$ as parameters and computes its output as:
\begin{lstlisting}
Y = conv_einsum("bmshw,nmhw,tshw->bnthw|hw",X,W1,W2).
\end{lstlisting}

% separable depth-wise convolution
{\bf (2)} In a {\em separable depth-wise convolution}, we assume the input and output channels are the same, i.e., $T = S$. 
The layer is parameterized by two matrices $\matrixSup{W}{1} \in \R^{S \times H}$ and $\mymatrix{W}_2 \in \R^{S \times W}$ such that
\begin{lstlisting}
Y = conv_einsum("bshw,sh,sw->bshw|hw", X, W1, W2)
\end{lstlisting}

We refer to \cite{hayashi2019exploring} for more examples.

\section{Optimal Sequencer}
\label{app:algorithms}

\opteinsum \cite{daniel2018opt} determines the optimal evaluation path of a tensor contraction sequence via the \netcon algorithm~\cite{pfeifer2014faster}. The algorithm considers the cost of evaluating intermediate products as it explores the path tree. We do not cover the tree traversal strategy here (we refer the reader to \cite{pfeifer2014faster}), but instead discuss how our \conveinsum generalizes \netcon by calculating the cost of a tensorial intermediate product. 
We first review the cost (i.e., number of additions/multiplications) of each primitive operation in FLOPs. Let $\tensorSup{T}{0}\in\mathbb{R}^{I_0 \times \cdots I_k \times I_{m-1}}$, $\tensorSup{T}{1}\in\mathbb{R}^{J_0 \times \cdots J_l  \cdots \times J_{n-1}}$:
\begin{enumerate}
    \item The mode-($k,l$) contraction cost is
    \begin{equation}
    \label{contraction-cost}
    \mathcal{O}\Bigl(\bigl(\prod_{p=0}^{m-1} I_p\bigr)\bigl(\prod_{q=0,q\neq l}^{n-1} J_q \bigr)\Bigr).
    \end{equation}
    \item The mode-($k,l$) batch product cost is 
    \begin{equation}
    \label{batch-cost}
    \mathcal{O}\Bigl(\bigl(\prod_{p=0}^{m-1} I_p\bigr)\bigl(\prod_{q=0, q\neq l}^{n-1} J_q \bigr)\Bigr).    
    \end{equation}
    \item The outer product cost is
    \begin{equation}
    \label{outer-prod-cost}
    \mathcal{O}\Bigl(\bigl(\prod_{p=0}^{m-1} I_p\bigr)\bigl(\prod_{q=0}^{n-1} J_q\bigr)\Bigr).
    \end{equation}
    \item The mode-($k,l$) convolution cost (without a Fast Fourier Transform) is
    \begin{equation}
    \label{convolution-cost}
    \mathcal{O}\Bigl(\bigl(\prod_{p=0}^{m-1} I_p\bigr)\bigl(\prod_{q=0}^{n-1} J_q\bigr)\Bigr).
    \end{equation}
\end{enumerate}
Now, as \netcon explores the path tree associated to contraction sequence (which includes batch products, outer products as special cases of contractions), it invokes a \textsf{cost} method which relies on \Cref{contraction-cost,outer-prod-cost} to analyze an intermediate tensor product along a path The \textsf{tnn-cost} method replaces the standard \textsf{cost} function of \netcon to fully realize the optimal sequencer by adding in the convolution cost model of \Cref{convolution-cost} (in addition to complex string handling to accommodate one letter of convolution type being associated to several dimensional sizes). 

% Discussion of convolution varieties
\textbf{Convolution Varieties.}
In our preceding discussions, we subtly assume commutative property of the convolution operation for optimal order evaluation. However, in practice, the convolutions used in neural networks are not necessarily commutative, since one input corresponds to features and another corresponds to filters. Specifically, if the convolution is not standard (e.g., dilated or strided) or not circularly padded, the convolution operation will not be communicative.
To make our \opentnn compatible with neural network practice, we support non-communicative convolutions if a letter for convolution only appears in two inputs. When non-communicative convolution is used, we assume the input with larger dimension size at the specified mode as features and another input as filters. 
However, if a letter for convolution appear more than twice in the inputs (i.e., the convolution is multi-way), we will only support communicative convolution with circular padding for now.

\textbf{Modification of the cost model for training.} 
Existing \einsum implementations only consider forward computation in tensor networks. However, in a neural network setting, we also need to consider the backpropagation computation. Further modification of \textsf{tnn-cost} is needed to incorporate backpropagation costs, which are once again tensorial sequences dictated by the same cost equations. In practice, we submit a flag to \conveinsum to indicate that the optimal sequencer is being used for training. 

Specifically, given two inputs $\tensorSup{T}{1}$, $\tensorSup{T}{2}$, which interact through an atomic operation $f$ resulting in an output tensor $\mytensor{T}=f(\tensorSup{T}{1},\tensorSup{T}{2})$, a standard \netcon-driven \einsum sequencer will calculate the cost of computing $\tensor{T}$ without any concern for associated backpropagation calculations. However, the backpropagation algorithm needs to compute ${\partial \mathcal{L}}/{\partial \tensorSup{T}{1}} = g_1({\partial \mathcal{L}}/ {\partial \mytensor{T}}, \tensorSup{T}{2})$ and ${\partial \mathcal{L}} / {\partial \tensorSup{T}{2}} = g_2(\tensorSup{T}{1}, {\partial \mathcal{L}} / {\partial \mytensor{T}})$, where $g_1$ and $g_2$ are gradient calculations dependent on $f$. Therefore, we modify the cost from $\textsf{cost}(f)$ to $\textsf{cost}(f) + \textsf{cost}(g_1) + \textsf{cost}(g_2)$. For instance, consider $f$ as a standard 2D-convolution, where the operation between the input $\tensorSup{T}{1} \in \R^{B \times S \times X \times Y}$ and the weight $\tensorSup{T}{2} \in \R^{T \times S \times H \times W}$ leads to the output $\tensor{T} \in \R^{B \times T \times X^\prime \times Y^\prime}$. We have $\textsf{cost}(f) = O(BHWXYTS)$ for the forward pass, and $\textsf{cost}(g_1) = O(B HW X^\prime Y^\prime TS)$, $\textsf{cost}(g_2) = O(B XY X^\prime Y^\prime TS)$ for the backward pass. In order to achieve optimal scheduling, we further modify the cost function of \netcon to consider all three such costs.
\label{app:theorems}
\section{Proof of Theorem \ref{flop-reduce} and \ref{flop-reduce-tucker}}
\begin{theorem*}[CP reduction]
\label{flop-reduce-full}
Let $\tensor{X}\in\mathbb{R}^{B \times S\times H' \times W'}$ be the input to a reshaped CP (RCP) convolutional kernel $\mytensor{\overline{W}} \in \R^{T_1 \cdots \times T_M \times S_1 \cdots \times S_M \times H \times W}$ such that $T = \prod_{m = 1}^{M} T_m$, $S = \prod_{m = 1}^{M} S_m$ are factored into $(m + 1)$ rank-$R$ CP factor tensors $\matrixSup{W}{m} \in \R^{R \times T_m \times S_m}$ with $\tensorSup{W}{0} \in \R^{R \times H \times W}$. Assume $H'\gg H$ and $W'\gg W$ are large; in particular $SH'W'>aHW$ and $BH'W'>aS$ for some constant $a\geq 1$. Furthermore, let $R\geq S$. Then the forward pass through the RCP kernel (in the syntax of \conveinsum), 
\begin{lstlisting}[mathescape=true]
Y=conv_einsum("b(s1)(s2)(s3)$\cdots$(sM)hw,r(t1)(s1),r(t2)(s2),r(t3)(s3),$\dots$, r(tM)(sM),rhw->r(t1)(t2)(t3)$\cdots$(tM)hw",X, W1, W2, W3, $\dots$, WM, W0)
\end{lstlisting}
has a pairwise evaluation path which costs less FLOPs than the following naive left-to-right evaluation,
 \begin{lstlisting}[mathescape=true]
Y=conv_einsum("br(t1)(t2)(t3)$\cdots$(tM)hw, rhw -> br(t1)(t2)(t3)$\cdots$(tM)hw| hw", YM, W0)
 \end{lstlisting}
 where
 \begin{lstlisting}[mathescape=true]
 Ym=einsum("r(s1)(s2)$\cdots$(s(m-1))(t1)(t2)$\cdots$(t(m-1)),r(sm)(tm)->r(s1)(s2)$\cdots$(sm)(t1)(t2)$\cdots (tm)$, Y(m-1), Wm) 
 \end{lstlisting}
 for $1\leq m \leq M$, noting that for tensor object \texttt{YM}, its mode symbols \texttt{h}, \texttt{w} correspond to dimensions $H',W'$, and \texttt{b} corresponds to an arbitrary batch size.
\end{theorem*}
\begin{proof}
 We first evaluate the number of multiplications and additions of the left-to-right evaluation. Let $U_k=\prod_{i=k}^MS_i \prod_{i=1}^k T_i$ for $1\leq k \leq M-1$ and $U_M=T$. According to the cost formulae and general definitions of the multilinear operations involved, the number of multiplications for the left-to-right evaluation is 
\begin{align*}
M_{\textrm{naive}}&=BRH'W'\sum_{i=1}^M U_i +BRTHWH'W'\\
&=BRH'W'(\sum_{i=1}^M U_i +THW).
\end{align*}
Additions are incurred by contractions over the $\texttt{s}_i$, and by the formula for a general multilinear contraction, in less frequency than the resulting multiplications, so to prove our claim, we only concern ourselves with demonstrating there is a path with less multiplications. 
%\begin{equation*}
%A_{\textrm{naive}}=BRH'W'\sum_{i=1}^MS_i,  
%\end{equation*}
%arising from sequential contractions over the $$\texttt{s}_i$.
Now, we will choose the following path,
 \begin{lstlisting}[mathescape=true]
Ym=conv_einsum("r(s1)(s2)$\cdots$(s(m-1))(t1)(t2)$\cdots$(t(m-1)),r(tm)(sm)->r(s1)(s2)$\cdots$(sm)(t1)(t2)$\cdots$(tm)", Y(m-1), Wm),
 \end{lstlisting}
 for $2\leq m \leq M$. We then perform 
 \begin{lstlisting}[mathescape=true]
Y0=conv_einsum("r(s1)(s2)$\cdots$(sm)(t1)(t2)$\cdots$(tm),rhw->(s1)(s2)$\cdots$(sm)(t1)(t2)$\cdots$(tm)hw, YM, W0)
 \end{lstlisting}
 followed lastly by
 \begin{lstlisting}[mathescape=true]
 Y=conv_einsum("(s1)(s2)$\cdots$(sm)(t1)(t2)$\cdots$(tm)hw,b(s1)(s2)$\cdots$(sm)hw->br(t1)(t2)$\cdots$(tm)hw|hw, Y0, X).   
 \end{lstlisting}
Let $V_k=\prod_{i=1}^k S_iT_i$. The number of multiplications of this path is
\begin{equation*}
    M_{\textrm{reduced}} = R\sum_{i=1}^M V_i+RSTHW+BSTHWH'W'.
\end{equation*}
Now, for the first term of $M_{\textrm{reduced}}$,
\begin{equation*}
    R\sum_{i=1}^M V_i <RS\sum_{i=1}^M\prod_{j=1}^iT_i <\frac{BRH'W'}{a}\sum_{i=1}^M U_i.
\end{equation*}
with the last inequality following from our assumption that and $BH'W'>aS$. For the second term we have that
\begin{align*}
    RSTHW &<\frac{RTH'W'}{a}\\
    &<\frac{(a-1)RTH'W'}{a}\\
    &< \frac{(a-1)RH'W'\sum_{i=1}^M U_i}{a}\\
    &<\frac{(a-1)BRH'W'\sum_{i=1}^M U_i}{a},
\end{align*}
where the second inequality follows from the fact $T<U_M$ (since $U_M$ contains T), therefore the entire series is larger than $T$. By the preceding two final inequalities we have then that
\begin{equation}
\label{first-terms}
    R\sum_{i=1}^M V_1+RSTHW < BRH'W'\sum_{i=1}^M U_i
\end{equation}
For the last term, by our assumption $R\geq S$, we have that
\begin{equation}
\label{last-term}
    BSTHWH'W' < BRTHWH'W'.
\end{equation}
Combining the upper bounds given by equations \ref{first-terms} and \ref{last-term}, we establish our result, $M_{\textrm{reduced}}<M_{\textrm{naive}}$. 

\textit{Remarks.} \textit{(1)} Our proof demonstrates the existence of a cheaper path. The FLOPs-minimal path is more complicated and depends on the sizes of the factor tensors $S_i, T_i$. \textit{(2)} Our assumption $H',W' \gg H,W$ is common in image processing.natural one -- image size often greatly exceed filter size. 
\end{proof}

\begin{theorem*}[Tucker reduction]
\label{flop-tucker-full}
Let $\tensor{X}\in\mathbb{R}^{B \times S\times H' \times W'}$ be the input to a reshaped Tucker (RTK) convolutional kernel $\mytensor{\overline{W}} \in \R^{T_1 \cdots \times T_M \times S_1 \cdots \times S_M \times H \times W}$ such that $T = \prod_{m = 1}^{M} T_m$, $S = \prod_{m = 1}^{M} S_m$ are factored into $(m + 1)$ rank-$R_m$ CP factor tensors $\matrixSup{W}{m} \in \R^{R_m \times T_m \times S_m}$ with $\tensorSup{W}{0} \in \R^{R \times H \times W}$ and $\tensorSup{C} \in \R^{R_0 \times R_1 \times \cdots \times R_M}$. Assume $H'\gg H$ and $W'\gg W$ are large; in particular $SH'W'>aHW$ and $BH'W'>aS$ for some constant $a\geq 1$. Furthermore, let $\prod_{i=1}^M R_m\geq S$. Then the forward pass through the RCT kernel (in the syntax of \conveinsum), 
\begin{lstlisting}[mathescape=true]
Y=conv_einsum("b(s1)(s2)(s3)$\cdots$(sM)hw,(r1)(t1)(s1),(r2)(t2)(s2),(r3)(t3)(s3),$\dots$, (rM)(tM)(sM),(r0)hw, (r0)(r1)(r2)$\cdots$(rM)->b(t1)(t2)(t3)$\cdots$(tM)hw|hw",X, W1, W2, W3, $\dots$, WM, W0, C)
\end{lstlisting}
has a pairwise evaluation path which costs less FLOPs than the naive left-to-right evaluation,
 \begin{lstlisting}[mathescape=true]
Y=conv_einsum("br(t1)(t2)(t3)$\cdots$(tM)hw, rhw -> br(t1)(t2)(t3)$\cdots$(tM)hw| hw", YM, W0)
 \end{lstlisting}
 where
 \begin{lstlisting}[mathescape=true]
 Ym=einsum("r(s1)(s2)$\cdots$(s(m-1))(t1)(t2)$\cdots$(t(m-1)),r(sm)(tm)->r(s1)(s2)$\cdots$(sm)(t1)(t2)$\cdots$ (tm), Y(m-1), Wm) 
 \end{lstlisting}
 for $1\leq m \leq M$, noting that for tensor object \texttt{YM}, its mode symbols \texttt{h}, \texttt{w} correspond to dimensions $H',W'$, and \texttt{b} corresponds to an arbitrary batch size.
\end{theorem*}
\begin{proof}
The proof follows a nearly identical argument as in the proof of previous Theorem with instead with the only difference being slightly different intermediate osts according to the $R_m$ modes of the $\tensorSup{W}^m$, which may all be upper-bounded by $R$.  
\end{proof}
\section{Supplementary Experimental Data}
\label{app:experiments}
In this section, we report on additional data relating to not included in the manuscript. 
\begin{table}[!htbp]
\caption{\textbf{Run-time (seconds per epoch) comparison between \conveinsum and \pytorch implementation of TNNs using different tensor decomposition forms in the image classification task on the CIFAR-10 dataset}. The base architecture used is ResNet-34. We observe that \conveinsum outperforms \pytorch under different tensor decompositions. Here, RXX denotes the reshaped XX decomposition, with reshaping factor M=3.}
 \label{tab:different-tensor-forms}
  \centering
  \begin{tabular}{lcccccc}
    \toprule
    & \multicolumn{2}{c}{conv-einsum} & \multicolumn{2}{c}{\shortstack{Naive \\w/o ckpt}} & \multicolumn{2}{c}{\shortstack{Naive\\w/ ckpt}} \\
    \cmidrule(lr){2-3}\cmidrule(lr){4-5}\cmidrule(lr){6-7}
    \thead{Tensor Form} & \thead{Train} & \thead{Test} & \thead{Train} & \thead{Test} & \thead{Train} & \thead{Test}\\
    \midrule
    RCP & 14 & 2.14 & 22 & 2.57 & 29 & 2.61\\
    RTR & 8 & 1.41 & 16 & 1.86 & 16 & 1.86\\
    RTT & 8 & 1.37 & 16 & 1.61 & 16 & 1.68\\
    RTK & 6 & 1.34 & 17 & 1.47 & 17 & 1.54\\
    \bottomrule
\end{tabular}
\end{table}
\begin{table}[!htbp]
\vspace{-1em}
\caption{\textbf{TNN performance under various model scales for image classification under low resources (4-core CPU).}} CIFAR10 classification using RCP ($M=3$) ResNet-34 and a batch size of 128. Result is seconds per epoch.\\
 \label{tab:runtime-cpu}
  \centering
\resizebox{0.6\linewidth}{!}{\begin{tabular}{lllll}
    \toprule
    \thead{Compression Rate (CR)} & \thead{RCP-train} & \thead{RCP-test}  & \thead{TK-train} & \thead{TK-test}\\
    \midrule
    100\%   & 2030 & 132 & 187  & 12.9\\
    50\%    & 1378 & 97  & 151  & 10.6\\
    20\%    & 1054 & 76   & 150   & 10.1\\
    10\%    & 872  & 65   & 160   & 9.5\\
    5\%     & 785  & 58   & 147   & 9.0\\
    \bottomrule
 \end{tabular}}
\end{table}
\begin{table}
\caption{\protect\textbf{TNN performance under various model scales for diverse machine learning tasks.} Automatic Speech Recognition (ASR) on LibriSpeech is measured by Word Error Rate (WER) (the lower, the better). Image classification (IC) on CIFAR10 (results from \protect\cite{su2018tensorial}) is measured by top-1 precision. Video classification (VC) on UCF-101 is measured by top-1 accuracy. Averaged over 3 runs.}
 \label{tab:performance-rcp}
  \centering
%\resizebox{0.5\columnwidth}{!}
{
\begin{tabular}{llll}
    \toprule
    \thead{Compression Rate (CR)} & \thead{IC} & \thead{ASR}  & \thead{VC}\\
    \midrule
    Original & 93.2  & 2.1 &  88.98\\
    \midrule
    100\%   & -     & 2.08 &  89.00\\
    50\%    & -     & 2.29 &  88.61\\
    20\%    & -     & 2.36 &  88.10\\
    10\%    &91.28  & 2.43 & 87.63\\
    5\%     &89.86  & 3.01 &  86.62\\
    2\%     &85.70  & 3.76 &  86.41\\
%    1\%     &\textcolor{red}{78.61}  & ? & 3.92 &  84.03\\
    \bottomrule
 \end{tabular}}
\end{table}

\end{document}